%% file: main.tex
\documentclass[10pt,twocolumn,letterpaper]{article}

\usepackage[pagenumbers]{iccv} % To force page numbers, e.g. for an arXiv version

\usepackage{multirow}
\usepackage{tabularx}


\definecolor{iccvblue}{rgb}{0.21,0.49,0.74}
\usepackage[pagebackref,breaklinks,colorlinks,allcolors=iccvblue]{hyperref}

\def\paperID{4395} % *** Enter the Paper ID here
\def\confName{ICCV}
\def\confYear{2025}

\title{XR-VLM: Cross-Relationship Modeling with Multi-part Prompts and Visual Features for Fine-Grained Recognition}

\author{Chuanming Wang\\
\and
Hengming Mao\\
\and
Huanhuan Zhang\\
\and
Huiyuan Fu\\
\and
Huadong Ma\\
\and 
The State Key Laboratory of Networking and Switching Technology\\
Beijing University of Posts and Telecommunications\\
{\tt\small \{wcm, maohengming, zhanghuanhuan, fhy, mhd\}@bupt.edu.cn}
}

\begin{document}
\maketitle

\input{sec/0_abstract}    
\input{sec/1_intro}
\input{sec/2_related}

\input{sec/3_method}
\input{sec/4_experiment}
\input{sec/5_conclusion}

{
    \small
    \bibliographystyle{ieeenat_fullname}
    \bibliography{main}
}
\input{sec/X_suppl.tex}

\end{document}

%% file: sec/0_abstract.tex
\begin{abstract}
    Vision-Language Models (VLMs) have demonstrated impressive performance on various visual tasks, yet they still require adaptation on downstream tasks to achieve optimal performance. Recently, various adaptation technologies have been proposed, but we observe they often underperform in fine-grained visual recognition, which requires models to capture subtle yet discriminative features to distinguish similar sub-categories. Current adaptation methods typically rely on an alignment-based prediction framework, \ie the visual feature is compared with each class prompt for similarity calculation as the final prediction, which lacks class interaction during the forward pass. Besides, learning single uni-modal feature further restricts the model's expressive capacity. Therefore, we propose a novel mechanism, XR-VLM, to discover subtle differences by modeling cross-relationships, which specifically excels in scenarios involving multiple features. Our method introduces a unified multi-part visual feature extraction module designed to seamlessly integrate with the diverse backbones inherent in VLMs. Additionally, we develop a multi-part prompt learning module to capture multi-perspective descriptions of sub-categories. To further enhance discriminative capability, we propose a cross relationship modeling pattern that combines visual feature with all class prompt features, enabling a deeper exploration of the relationships between these two modalities. Extensive experiments have been conducted on various fine-grained datasets, and the results demonstrate that our method achieves significant improvements compared to current state-of-the-art approaches.
    Code will be released.
\end{abstract}

%% file: sec/1_intro.tex
\vspace{-0.5em}
\section{Introduction}
\label{sec:intro}
Vision-Language Models (VLMs)~\cite{DBLP:conf/emnlp/XuG0OAMZF21, DBLP:conf/icml/RadfordKHRGASAM21, DBLP:conf/icml/JiaYXCPPLSLD21, DBLP:conf/cvpr/ZhaiWMSK0B22, DBLP:conf/iclr/YaoHHLNXLLJX22} have achieved significant success by using contrastive learning to project images and texts into a shared feature space during pre-training, which has encouraged researchers to adapt these models for various downstream tasks. However, due to the need of processing both vision and language data, VLMs typically contain a huge number of parameters, making them substantially more resource-intensive than traditional backbones~\cite{DBLP:journals/corr/SimonyanZ14a, DBLP:conf/cvpr/HeZRS16}. As a result, fine-tuning all parameters in these models is not practical due to its massive requirements of considerable data and computational resources. To address these challenges, a growing number of studies~\cite{DBLP:journals/ijcv/ZhouYLL22, DBLP:conf/iclr/HuSWALWWC22, DBLP:conf/cvpr/YangZWX24} have focused on optimizing specific modules within VLMs rather than fine-tuning the entire model. In the literature, prompt learning and adapter learning have emerged as promising paradigms, recognized for their flexibility and efficiency in enhancing VLM performance across various downstream tasks.

\begin{figure}[t]
\centering
	\begin{subfigure}{0.492\linewidth}
		\includegraphics[width=\textwidth]{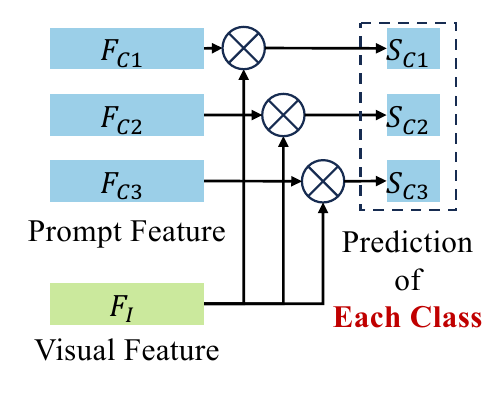}
		\caption{Illustration of Aligning Pattern.}
		\vspace{-0.5em}
		\label{fig:intro-a}
	\end{subfigure}
\hfill
	\begin{subfigure}{0.492\linewidth}
		\includegraphics[width=\textwidth]{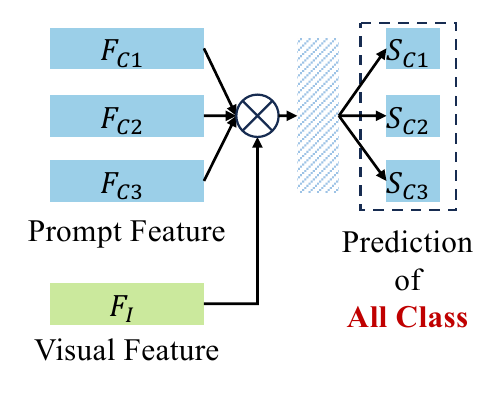}        
		\caption{Illustration of Crossing Pattern.}
		\vspace{-0.5em}
		\label{fig:intro-b}
	\end{subfigure}
\vspace{-1.2em}
\caption{Comparison between (a) previous prediction pattern, generated by aligning a single visual feature with each class prompt feature individually (referred to as the \textit{aligning pattern}), and (b) our prediction pattern, generated by modeling cross relationship between a single visual feature and all class prompt features collectively (referred to as the \textit{crossing pattern})}
\label{fig:intro}
\vspace{-1.5em}
\end{figure}

Prompt learning~\cite{DBLP:journals/ijcv/ZhouYLL22, DBLP:conf/cvpr/DuWZSGL22, DBLP:conf/iccv/ZhuLCFMJLS23} focuses on the optimization of language input by designing task-specific textual prompts that guide the model to generate more relevant and context-aware features. On the other hand, adapter learning methods insert lightweight, task-specific modules into the pre-trained model, enabling improve the discriminativeness of outputs from both the vision and language branches. Despite differences in the parameters being fine-tuned, these methods still rely on an alignment-based pattern for prediction. As illustrated in Fig.~\ref{fig:intro-a}, the visual feature is compared with each class prompt feature, and the resulting similarity is used as the logit for the corresponding class. However, we find that such a prediction pattern is not well-suited for fine-grained visual recognition (FGVR)~\cite{DBLP:conf/eccv/ZhangDGD14}, a task that requires distinguishing highly similar sub-categories within a broader class (e.g., differentiating bird species or car models). The similarity among these sub-categories is significantly higher compared to coarse-grained classes (a simple evaluation experiment demonstrating this is provided in the supplementary material), often leading to confusion in predictions. Furthermore, previous methods typically output a single feature for both vision and language branches, which restricts the model's expressive capacity.

To address this issue, we argue that modeling relationships is crucial for FGVR and the final prediction should be taken by considering all class information. Therefore, we propose a \underline{Cross}-\underline{R}elationship modeling method (XR-VLM), that captures interactions across prompts and visual features. As shown in Fig.~\ref{fig:intro-b}, the key distinction between our and previous methods is that we aggregate the similarities between a visual feature and all class prompts into a relationship representation, which is then used for predicting the logits of all classes. It can be seen as shifting from a 1-to-1 alignment to a 1-to-many comparison scheme. As a result, the final prediction $P_i$ of class $i$ incorporates information from all other classes, enabling richer contextual reasoning. Additionally, we introduce a multi-part prompt and visual feature learning strategy to enhance the expressiveness of learned features, further supporting the modeling of cross relationships.

To evaluate our method, we conduct extensive experiments on various fine-grained benchmark datasets, comparing it with state-of-the-art (SoTA) VLM adaptation methods. Experimental results demonstrate that our approach achieves significant improvements over SoTA methods and attains the best performance on nearly all datasets. Extensive quantitative and visual analyses further confirm that modeling class relationships leads to better performance compared to traditional alignment-based patterns, and learning multiple prompts and visual featuers per class further enhances VLMs' effectiveness on FGVR.

The contributions can be summaried as follows:
\begin{itemize}
	\item We propose a multi-part prompt and visual feature learning framework for VLMs, designed to fully harness the complementary strengths of both vision and language information, enabling more precise capture of subtle differences across similar classes.

	\item We introduce a cross-relationship modeling scheme to replace the traditional alignment-based prediction pattern. It enables richer interactions among classes and allow the model to incorporate contextual information from other classes during prediction.

	\item Extensive experiments on fine-grained benchmarks demonstrate consistent improvements over state-of-the-art methods, validating the effectiveness of our method.
\end{itemize}

%% file: sec/2_related.tex
\section{Related Works}
\subsection{Vision-Language Models}
Foundational Vision-Language Models (VLMs) leverage both vision and language modalities to learn rich multi-modal representations. These models utilize a self-supervised paradigm, trained on large-scale image-text pairs from the web, enabling impressive performance across various downstream applications. However, efficiently adapting these pre-trained models to specific tasks remains a significant challenge. Conventional fine-tuning methods, which involve optimizing all parameters of the pre-trained network, are impractical due to the large number of parameters and the high computational resources. As a solution, prompt learning~\cite{DBLP:journals/ijcv/ZhouYLL22, DBLP:conf/cvpr/ZhouYL022, DBLP:conf/cvpr/KhattakR0KK23, DBLP:conf/iccv/KhattakWNK0K23, DBLP:conf/cvpr/LiLFZWCY24} has been proposed, which aims at learning suitable addtional context on the input text data. Existing work has primarily focused on how to learn prompts for downstream tasks. For example, PromptSRC~\cite{DBLP:conf/iccv/KhattakWNK0K23} uses original features to regularize the prompt learning process; PromptKD~\cite{DBLP:conf/cvpr/LiLFZWCY24} interpolates a distillation framework to transfer knowledge from a large teacher network to a small student network by prompts. There are some works learn both text and visual prompts, \eg MaPLE~\cite{DBLP:conf/cvpr/KhattakR0KK23} learns prompts for both the image and text branches simultaneously. Adapter learning is another paradigm for VLM adaptation, which focus on integrate additional modules to the pre-trained model and only fine-tuning such modules to boost the discriminativeness of output features. For example, CLIP-adapter~\cite{DBLP:journals/ijcv/GaoGZMFZLQ24} adopts an additional bottleneck layer to learn residual feature blending with the original outputs, MMA~\cite{DBLP:conf/cvpr/YangZWX24} projects vision and language features into a share space for fine-tuning, Tip-Adapter-F~\cite{DBLP:conf/eccv/ZhangZFGLDQL22} adds a nonlinear, quadratic-complexity module and blends the class scores with the original textual features by evaluating the pairwise similarities between the features of the support sets. 
However, these methods typically use a single prompt per class and rely on conventional 1-to-1 alignment for prediction, which is not ideal for FGVR. In contrast, we propose a framework that utilizes cross-relationship representations to improve the prediction of similar classes in FGVR task.

\subsection{Fine-grained Visual Recognition}
FGVR aims to distinguish different sub-categories from one general super-class, which is crucial for many real-world applications. As a result, significant progress has been made in this field over the past few decades. The related literature can be broadly divided into two groups: part-based~\cite{DBLP:conf/eccv/ZhangDGD14, DBLP:conf/cvpr/FuZM17, DBLP:conf/iccv/Rao0L021} and part-free methods~\cite{DBLP:conf/iccv/LinRM15, DBLP:conf/cvpr/ZhengFZL19, DBLP:conf/cvpr/YangWCX022}. Part-based methods are driven by the need to differentiate similar sub-categories by capturing subtle differences in discriminative parts. They focus on learning to locate and emphasize various part regions using detection~\cite{DBLP:conf/nips/RenHGS15} or attention~\cite{Hu2017SqueezeandExcitationN} sub-networks, followed by feature extraction for classification. In contrast, part-free methods enhance the backbone model's recognition ability by re-weighting features~\cite{DBLP:conf/cvpr/WangMD18}, high-order pooling techniques~\cite{DBLP:conf/iccv/LinRM15}, or employing augmented samples~\cite{DBLP:conf/cvpr/ZhengFZL19} during training. Both types of methods follow the traditional classification paradigm, where an input image is processed by a visual backbone to extract deep features, which are then mapped to a predefined class space using a Multilayer Perceptron (MLP). However, VLMs are trained under fundamentally different settings, making direct comparisons with these methods challenging. Therefore, to ensure a fair comparison, we exclude these methods from our experiments.

\subsection{Multiple Feature Learning}
Multiple feature learning has been widely used in computer vision fiels. For example, MCTrans~\cite{DBLP:conf/cvpr/XuOBB022} embeds multiple class tokens into the Vision Transformer (ViT) backbone, and it obtains semantic segmentation results of input images by calculating the similarity between different class tokens and patch tokens. AP-Net~\cite{DBLP:journals/tip/ChenGLBZ21} adopts an attention pyramid module to obtain multiple features progressively, and it combines them to calculate similarities between query and gallery images. Specially, learning multiple features is an important mechanism that can boost models' performance on various FGVR tasks. For example, WS-DAN~\cite{DBLP:journals/corr/abs-1901-09891}, CAL~\cite{DBLP:conf/iccv/Rao0L021}, and MEPR~\cite{DBLP:journals/tmm/WangFM24} design different attention module to produce multiple  part features for classification. These methods focus on how to ensure the diversity of extracted features by different loss functions or attention generation operations. While our method also involves learning multi-part features, it uniquely emphasizes the learning and utilization of multiple textual prompts. 
A related work is PLOT~\cite{DBLP:conf/iclr/0002YSLR023}, which also employs multiple prompts and aligns them with local features using an optimal transport algorithm. In contrast, we introduce a multi-part feature extraction module for visual data, enriching feature semantics for prediction. More importantly, PLOT still adheres to the aligning pattern for prediction, which fundamentally differs from our method.

%% file: sec/3_method.tex
\section{Method}
In this section, we begin by introducing the fundamental processes of VLMs and prompt learning. Next, we briefly provide an overview of our proposed XR-VLM, demonstrating how multi-part prompts and visual features are utilized into the framework. Finally, we detail the implementation of each module in our method.

\begin{figure*}[t]
	\centering
	\includegraphics[width=0.83\linewidth]{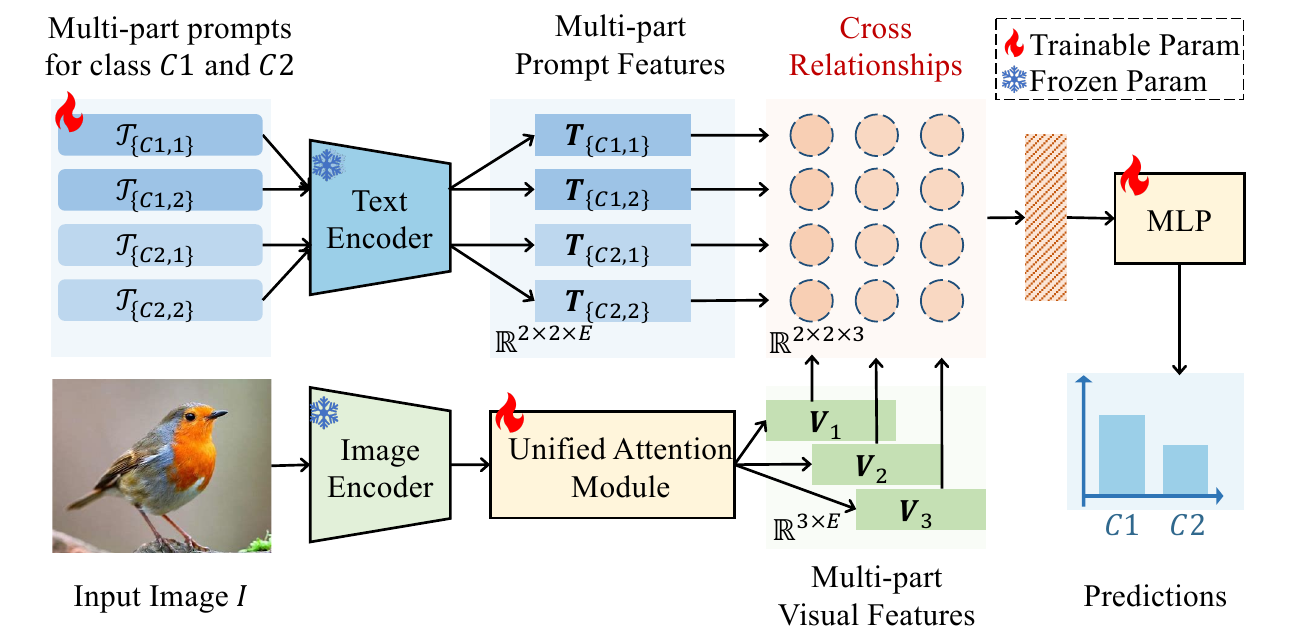}
	\vspace{-1em}
	\caption{Overall framework of our proposed XR-VLM: For the text branch, multi-part learnable prompts of classes are fed into the Text Encoder, generating multi-part prompt features. For the image branch, the input image is processed through the Image Encoder and an Unified Attention module to generate multi-part visual features. These prompt and visual features are sent to generated the cross-relationship representations, which is finally sent to a MLP module to generate the predictions. (best view in color) }
	\label{fig:framework}
	\vspace{-1em}
\end{figure*}

\subsection{Preliminary}
\textbf{Vision-Language Models.}
Following previous works~\cite{DBLP:journals/ijcv/ZhouYLL22, DBLP:conf/iclr/0002YSLR023, DBLP:conf/cvpr/HuangSDBBA24}, we adopt CLIP as our foundation model, which comprises two branches: an image encoder and a text encoder. Denoting the image encoder as $f^I$ and the text encoder as $f^T$, contrastive learning is applied between the input image $x$ and their corresponding textual descriptions $D^T=\{t_1, t_2, \dots, t_N\}$ in a prediction form. 

\begin{equation}
\begin{aligned}
    \label{eq:vlm}
	\mathrm{logits}\left(y=i|x\right) = \mu\left(f^I(x)\right)\cdot \mu\left(f^T(t_i)\right)^\top
\end{aligned}
\end{equation}
where $\mu$ is a normalization function for a vector defined as $\mu(x) = \frac{x}{||x||_2}$. Then, the output logit is used to calculate the cross-entropy loss to optimize both encoders. This process is conducted during the pre-training phase, and subsequent works have adopted a same operation in adapting to downstream tasks. Although the $\operatorname{softmax}$ function is used to map logits into a probability distribution for loss calculation—where class logits influence each other—this interaction occurs too late in the process. Thus, the model's ability to capture nuanced relationships between classes during the forward pass remains limited, which is particularly critical for FGVR.

\noindent \textbf{Prompt Learning.} 
In the original CLIP model, the textual descriptions are collected from the web and fixed during pre-training. Rather than relying on manually crafted prompts, recent work has proposed adaptively learning soft textual prompts for downstream tasks. Specifically, some learnable vectors are concatenated with the class name to create a contextualized representation. In this approach, the prompt for a class is represented as $t_i = {[V_1][V_2] \cdots [V_M][C_k]}$, where each vector $[V_j]$ is a learnable parameter and it has the same dimension as the word embeddings of class name $[C_k]$.

\subsection{Framework}
The overall framework of XR-VLM is illustrated in Fig.~\ref{fig:framework}. Multi-part prompts are fed into the text encoder to generate multi-part prompt features, while the input image is processed by the image encoder and transformed into multi-part visual features through an unified attention module. Subsequently, cross-relationship modeling is applied to these features to produce a cross-relationship representation, which is used for final prediction by a MLP module.

\subsection{Multi-part Prompt Learning}
Unlike previous prompt learning methods that typically use a single prompt per class, we pre-define multiple prompts for each class, each representing a different part description. The set of multi-part prompts for a given class, which are fed into the text encoder, is defined as:

\begin{equation}
    \mathcal{T}_k = \bigcup_i^S\left\{[V^i_1][V^i_2]\cdots[V^i_M][C_k] \right\},
\end{equation}
where $[V^s_m] \in \mathbb{R}^{E}$ ($s \in {1, \dots, S}$, $m \in {1, \dots, M}$) represents a vector with dimension $E$, matching the size of word embeddings, $M$ denotes the number of context of each part prompts, $S$ is the number of part prompts for each class, and $[C_k]$ is the word embedding of the $k$-th class name. The operation $\bigcup$ refers to the union of sets. As described in~\cite{DBLP:journals/ijcv/ZhouYLL22}, in $\mathcal{T}_k$, all $[V_j^i]$ are learnable, meaning they can be optimized via backward gradients. These pre-defined multi-part prompts are fed into the text encoder to produce the text features $\mathbf{T} \in \mathbb{R}^{W \times S \times E}$, where $W$ is the number of classes and $E$ is the feature dimension. Note that we do not share prompts across classes, as we find that doing so negatively impacts performance as the results reported in~\cite{DBLP:journals/ijcv/ZhouYLL22}. Instead, each class has its unique set of multi-part prompts.

\subsection{Multi-part Visual Feature Learning}
\label{sec:uni-atten}
Denoted the deep features learned by the visual encoder (e.g. ResNet~\cite{DBLP:conf/cvpr/HeZRS16} or ViT-B/16~\cite{DBLP:conf/iclr/DosovitskiyB0WZ21}) as $\mathbf{X}\in \mathbb{R}^{N\times E}$ (details is described in supplementary material), where $N$ and $E$ mean the number and channel dimension of features, respectively, we first adopt an effective \textit{Unified Attention} (UnA) module to generate multi-part visual features according to the attentions generated by:
\begin{equation}
\label{eq:atte}
    \mathbf{A} = \phi(\mathbf{X}; \mathbf{W}_\phi) \in \mathbb{R}^{N \times S}.
\end{equation}
In detail, $\phi$ is made up of a BatchNorm layer, a Linear layer, and a ReLU layer. The output dimension of the Linear layer is set to $S$, so there will be $S$ visual part features as a result, which is same with the number of prompts for each class, and each attention will respond to a particular pattern of object parts. Inspired by~\cite{DBLP:journals/tmm/WangFM24}, to make the learned attentions diverse, we adopt a redundant strategy for $\mathbf{A}$. First, we change the output dimension of $\phi$ to $S+1$, and then we apply a softmax operation along the second dimension of $\mathbf{A}$:
\begin{equation}
\label{eq:softmax}
    \bar{\mathbf{A}}_{i, s} = \frac{\exp(\mathbf{A}_{i, s} )}{\sum_{s=1}^{S+1}\exp(\mathbf{A}_{i, s} )}.
\end{equation}
Next, we select the first $S$ attention maps from $\bar{\mathbf{A}}$ and combine it with $\mathbf{X}$ to produce multiple part features:
\begin{equation}
\label{eq:pool}
    \mathbf{V} = \mathbf{\bar{A}}^\top \otimes \psi\left(\mathbf{X};\mathbf{W}_\psi\right) \in \mathbb{R}^{S\times E}. 
\end{equation}
$\psi$ is a linear layer with parameter $\mathbf{W}_\psi$, which projects the visual feature to the relationship learning space, and $\otimes$ denotes the matrix multiplication.
Through this way, the first $S$ attention will focus on different parts of object while the remained attention will concentrate on the background, and discriminative information will be brought into part features. Finally, we apply a $\ell2$ normalization on part features:
\begin{equation}
    \mathbf{V}_{i, j} = \frac{\mathbf{V}_{i, j}}{\sum_{s=1}^{S} \sum_{e=1}^{E} |\mathbf{V}_{s, e}|^2} * \tau,
\end{equation}
where $\tau$ is a scale factor, and we set it to $64.0$ as default.

\subsection{Cross Relationship Modeling}
By incorporating the above modules, we obtain multiple textual features from multi-part prompts and multiple visual features from a single input image. The key challenge lies in effectively utilizing the rich information contained in these features to achieve precise classification. An alternative strategy is to compute the distance between corresponding visual and textual features for each part type and aggregate the results as the final prediction (called as Part-wise Cosine Similarity, PwCS), which can be defined as:
\begin{equation}
    \mathbf{\hat{y}}^{\mathrm{PwCS}} = \frac{1}{S} \sum_{s=1}^S \mu\left(\mathbf{V}_{s}\right) \otimes \mu\left(\mathbf{T}_{\cdot, s}\right) \in \mathbb{R}^{W}.
	\label{eq:pwcs}
\end{equation}
However, as discussed in the introduction, this 1-to-1 distance calculation scheme has limited the expression of VLMs for FGVR. Therefore, we propose a termed \textit{1-to-many relationship learning} scheme, \ie modeling cross relationship to explore the differences and similarities between the learned representations of multi-part prompts and visuals for prediction, termed as \textit{Cross Relationship Modeling} (CRM) module in our method. 
We re-arrange the shape of $\mathbf{T}$ to $E\times S \times W$ via a transpose operation. The cross-relationship matrix $\mathbf{R}$ of each visual data is calculated as:
\begin{equation}
	\begin{aligned}
		\label{eq:ips-sim}
		\mathbf{R} = \epsilon(\mathbf{V} \otimes \mathbf{T}) \in \mathbb{R}^{SSW},
	\end{aligned}
\end{equation}
where $\epsilon$ is a flatten operation. Then, we treat it as a representation of visual data for classification by a classifier $\vartheta$:
\begin{equation}
	\begin{aligned}
	\label{eq:pred-ip}
		\mathbf{\hat{y}}^{\mathrm{CRM}} =\vartheta(\mathbf{R}; W_{\vartheta}) \in \mathbb{R}^W,
	\end{aligned}
\end{equation}
where $\vartheta$ is the classifier that consists of an MLP (FC-BN-ReLU-FC). This classifier $\vartheta$ can been seen as one type of parameterized metric function, which measure the similarity between visual data and multi-part prompts of all class in an implicit manner. 

\begin{figure}[t]
	\centering
	\includegraphics[width=\linewidth]{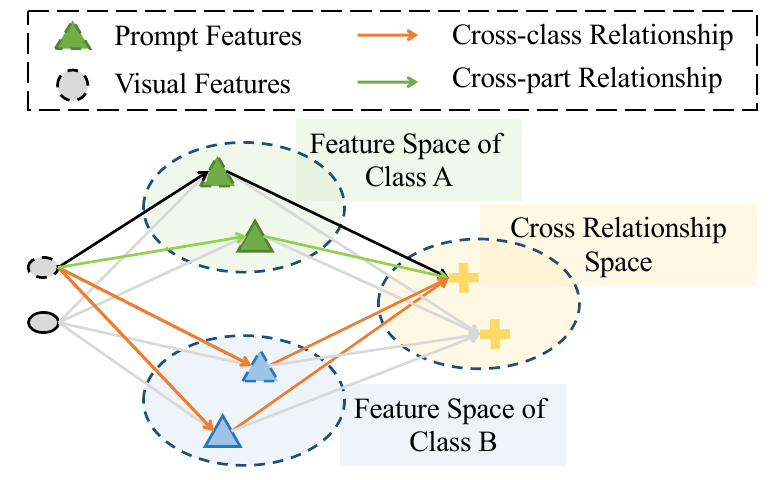}
	\caption{Illustration of cross relationships. Different shape borders represent different parts. (best view in color) }
	\label{fig:cross-part}
\end{figure}

In Fig.~\ref{fig:cross-part}, we give an illustration of cross relaitonship. Taking the upper visual feature as an example, given the feature spaces of class $A$ and class $B$, our method not only considers part-wise similarities (black lines) but also incorporates cross-part (green lines) and cross-class (orange lines) relationships. These relationships are projected into a unified space, where optimizing the distances within this space simultaneously influences all relationships. By comprehensively considering both similarities and differences across all classes, the model can capture subtle yet discriminative features, significantly enhancing its ability to distinguish fine-grained sub-categories.

%% file: sec/4_experiment.tex
\section{Experiments}

\begin{table*}[t]
	\centering
	\renewcommand\arraystretch{1.0}{
		\setlength{\tabcolsep}{0.7mm}{
			\begin{tabular}{clccccc|c}
				\specialrule{0.1em}{1pt}{1pt}  
				\hline
				\textbf{\begin{tabular}[c]{@{}c@{}}Visual\\ Encoder\end{tabular}} &
				\textbf{Methods} &
				\textbf{CUB-200-2011} &
				\textbf{Stanford-Cars} &
				\textbf{Stanford-Dogs} &
				\textbf{FGVC-Aircraft} &
				\textbf{NABirds} &
				\textbf{AVG.} \\
				\toprule

				\multirow{9}{*}{RN50}      
										& ZS-CLIP~\cite{DBLP:conf/icml/RadfordKHRGASAM21}$_{\mathrm{ICML'21}}$       & 45.3 & 55.7 & 52.3 & 17.2 & 36.5 & 41.4 \\
										& CLIP-Adapter~\cite{DBLP:journals/ijcv/GaoGZMFZLQ24}$_{\mathrm{IJCV'22}}$  & 55.8 & 63.5 & 58.5 & 21.9 & 45.5 & 49.0 \\
										& LP-CLIP~\cite{DBLP:conf/icml/RadfordKHRGASAM21}$_{\mathrm{ICML'21}}$       & 64.1 & 70.0 & 52.9 & 36.0 & 50.1 & 54.6 \\
                                        & ProGrad~\cite{DBLP:conf/iccv/ZhuNHWZ23}$_{\mathrm{ICCV'23}}$       & 64.2 & 73.3 & 65.6 & 30.3 & 46.2 & 55.9 \\
										& PLOT~\cite{DBLP:conf/iclr/0002YSLR023}$_{\mathrm{ICLR'23}}$          & 67.3 & 73.0 & 65.2 & 30.4 & 45.6 & 56.3 \\
                                        & CoOp~\cite{DBLP:journals/ijcv/ZhouYLL22}$_{\mathrm{IJCV'22}}$          & 67.0 & 73.0 & 63.9 & 31.4 & 47.8 & 56.6 \\
										& LP++~\cite{DBLP:conf/cvpr/HuangSDBBA24}$_{\mathrm{CVPR'24}}$          & 67.3 & 72.7 & 62.1 & 31.6 & 51.4 & 57.0 \\
										& Tip-Adapter-F~\cite{DBLP:conf/eccv/ZhangZFGLDQL22}$_{\mathrm{ECCV'22}}$ & 69.9 & 74.3 & 63.6 & 35.2 & 56.2 & 59.8 \\
                                        & XR-VLM(Ours)          & \textbf{75.7} & \textbf{81.8} & \textbf{68.8} & \textbf{53.5} & \textbf{67.7} & \textbf{69.5} \\ \midrule
				\multirow{11}{*}{ViT-B/16} 
										& ZS-CLIP~\cite{DBLP:conf/icml/RadfordKHRGASAM21}$_{\mathrm{ICML'21}}$       & 55.0 & 65.3 & 61.2 & 24.7 & 44.2 & 50.1 \\
										& MaPLE~\cite{DBLP:conf/cvpr/KhattakR0KK23}$_{\mathrm{CVPR'23}}$         & 68.2 & 73.9 & 73.9 & 36.9 & 55.9 & 61.8 \\
										& ProGrad~\cite{DBLP:conf/iccv/ZhuNHWZ23}$_{\mathrm{ICCV'23}}$       & 73.7 & 81.9 & 74.1 & 41.1 & 59.9 & 66.1 \\
                                        & LP++~\cite{DBLP:conf/cvpr/HuangSDBBA24}$_{\mathrm{CVPR'24}}$          & 74.9 & 80.9 & 72.7 & 41.7 & 60.7 & 66.2 \\
                                        & LP-CLIP~\cite{DBLP:conf/icml/RadfordKHRGASAM21}$_{\mathrm{ICML'21}}$       & 76.9 & 80.6 & 65.5 & 45.6 & 64.9 & 66.7 \\
                                        & CoOp~\cite{DBLP:journals/ijcv/ZhouYLL22}$_{\mathrm{IJCV'22}}$          & 76.7 & 82.2 & 74.2 & 43.4 & 61.4 & 67.6 \\
                                        & PromptSRC~\cite{DBLP:conf/iccv/KhattakWNK0K23}$_{\mathrm{ICCV'23}}$ & 75.3 & 80.5 & 76.9 & 44.2 & 61.6 & 68.1 \\
                                        & MMA~\cite{DBLP:conf/cvpr/YangZWX24}$_{\mathrm{CVPR'24}}$           & 76.3 & 80.8 & 76.9 & 43.4 & 64.1 & 68.3 \\
										& Tip-Adapter-F~\cite{DBLP:conf/eccv/ZhangZFGLDQL22}$_{\mathrm{ECCV'22}}$ & 79.1 & 83.7 & 73.5 & 44.6 & 68.2 & 69.8 \\
										& PLOT~\cite{DBLP:conf/iclr/0002YSLR023}$_{\mathrm{ICLR'23}}$          & 77.6 & 84.4 & \textbf{77.0} & 46.7 & 64.3 & 70.0 \\
										& XR-VLM(Ours)          & \textbf{81.1} & \textbf{86.5} & 75.2 & \textbf{57.4} & \textbf{74.1} & \textbf{74.9} \\ 
										\bottomrule \specialrule{0.1em}{1pt}{1pt}
			\end{tabular}
		}
	}
    \vspace{-1em}
	\caption{Comparison results for fine-grained visual recognition. The highest accuracy is marked as bold.}~\label{tab:sota}
    \vspace{-1.8em}
	\end{table*}
	
\subsection{Experimential Settings}
\textbf{Datasets.} 
Experiments are conducted on five widely used fine-grained visual recognition benchmark datasets: CUB-200-2011~\cite{WelinderEtal2010}, Stanford-Cars~\cite{DBLP:conf/iccvw/Krause0DF13}, Stanford-Dogs~\cite{KhoslaYaoJayadevaprakashFeiFei_FGVC2011}, FGVC-Aircraft~\cite{maji13fine-grained}, and NABirds~\cite{DBLP:conf/cvpr/HornBFHBIPB15}. The training and test sets are constructed following the settings established in prior works~\cite{DBLP:journals/ijcv/ZhouYLL22, DBLP:conf/cvpr/ZhouYL022, DBLP:conf/cvpr/LiLFZWCY24, DBLP:conf/cvpr/YangZWX24}.

\noindent \textbf{Optimize Configures.} CLIP~\cite{DBLP:conf/icml/RadfordKHRGASAM21} is adopted as the foundational VLM for experiments and we keep its image encoder and text encoder fixed during training. Only parameters in multi-part prompts, UnA module and MLP are used for optimization. We use SGD as the optimizer and set its weight-decay to 1e-4. We train the models with 100 epochs for all experiments. The learning rate is set to 2e-3 and is adjusted by cosine learning rate schedule. As the defaults, $M$ is set to 16 and $S$ is set to 4. Following previous methods, for each class, $16$ samples are randomly selected from the training set and the model is evaluated on all testing samples.

\subsection{Comparison with SoTAs}

We compare the performance of XR-VLM against a range of state-of-the-art methods: ZS-CLIP~\cite{DBLP:conf/icml/RadfordKHRGASAM21}, CLIP-Adapter~\cite{DBLP:journals/ijcv/GaoGZMFZLQ24}, LP-CLIP~\cite{DBLP:conf/icml/RadfordKHRGASAM21}, CoOp~\cite{DBLP:journals/ijcv/ZhouYLL22}, PLOT~\cite{DBLP:conf/iclr/0002YSLR023}, MaPLE~\cite{DBLP:conf/cvpr/KhattakR0KK23}, MMA~\cite{DBLP:conf/cvpr/YangZWX24}, LP++~\cite{DBLP:conf/cvpr/HuangSDBBA24}, ProGrad~\cite{DBLP:conf/iccv/ZhuNHWZ23}, PromptSRC~\cite{DBLP:conf/iccv/KhattakWNK0K23}, and Tip-Adapter-F~\cite{DBLP:conf/eccv/ZhangZFGLDQL22}. Notably, some methods, such as MaPLE and MMA, are specifically designed for a single type of encoder (e.g., Vision Transformers), and thus their results are only reported for that specific architecture. 

The experimental results are presented in Table~\ref{tab:sota}, and the methods are sorted by \textit{AVG.} column. From the results, it is evident that our proposed method achieves significant improvements over all compared baselines across most benchmark datasets. Specifically, XR-VLM outperforms existing methods on both RN50 and ViT-B/16 encoders, achieving the highest accuracy on CUB-200-2011, Stanford-Cars, Stanford-Dogs, FGVC-Aircraft, and NABirds datasets.
For instance, on the RN50 encoder, XR-VLM achieves an average accuracy of 69.5\%, surpassing the second-best method (Tip-Adapter-F) by a notable margin. Similarly, on the ViT-B/16 encoder, XR-VLM attains an average accuracy of 74.9\%, further solidifying its superiority.

Prompt learning methods, such as CoOp, ProGrad, and MaPLE, improve over zero-shot CLIP but often struggle to capture fine-grained details due to their reliance on single-prompt and aligning pattern. In contrast, XR-VLM's multi-part prompt and cross-relationship modeling address this limitation by enabling richer feature interactions. Adapter-based methods, such as Tip-Adapter-F, show competitive performance but are constrained by their alignment-based prediction patterns, which limit their ability to model complex relationships between classes. XR-VLM's cross-relationship modeling provides a more comprehensive approach to feature interaction, resulting in superior performance. Linear probing methods, such as LP-CLIP and LP++, while simple, lack the flexibility to model fine-grained relationships, leading to lower accuracy compared to XR-VLM. In general, our method achieves the highest accuracy on most tasks, and the consistent improvements over state-of-the-art baselines underscore the importance of cross-relationship modeling and multi-part feature learning in addressing the challenges of FGVR.

\subsection{Albation Study}
To fully explore the reasons for the superior performance of the proposed method and inspire the follow-up research work, we take a comprehensive ablation study. The experiments are conducted on CUB-200-2011 and Stanford-Cars datasets, two standard fine-grained image analysis datasets. The former are natural creatures, usually named by biological experts, whose names reflect some of their own attributes (Black footed Albatross, White necked Raven), while the latter are a typical class of artificial creatures, whose names contain fewer categorical attributes (Audi S4 Sedan 2012, BMW M5 Sedan 2010).

\begin{figure}[t]
	\centering
	\includegraphics[width=\linewidth]{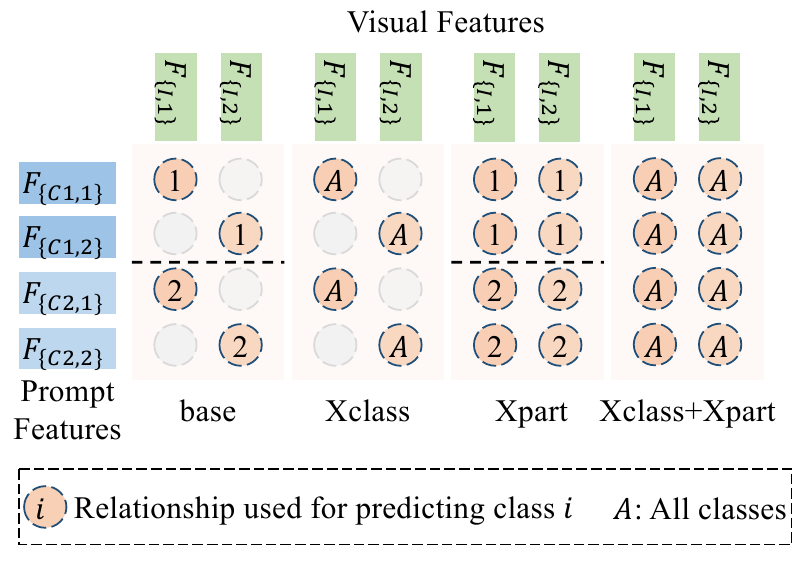}
    \vspace{-1.5em}
	\caption{Illustration of different cross relationships for prediction.}
	\label{fig:cross}
    \vspace{-1em}
\end{figure}

\begin{table}[]
\centering
\renewcommand\arraystretch{1.0}{
    \setlength{\tabcolsep}{1.8mm}{
        \begin{tabular}{clcccc}
            \specialrule{0.1em}{1pt}{1pt} 
            \hline
            \multirow{2}{*}{\textbf{\begin{tabular}[c]{@{}c@{}}Visual\\ Encoder\end{tabular}}} &
              \multicolumn{1}{c}{\multirow{2}{*}{\textbf{Strategy}}} &
              \multicolumn{4}{c}{\textbf{Datasets}} \\ \cline{3-6} 
             &
              \multicolumn{1}{c}{} &
              \multicolumn{2}{c}{\textbf{CUB-200-2011}} &
              \multicolumn{2}{c}{\textbf{Stanford-Cars}} \\ \hline
            \multirow{7}{*}{\rotatebox{90}{RN50}}   
                    & PLOT        & 67.31 & ${\mathrm{Ref.}}$         & 72.99 & ${\mathrm{Ref.}}$          \\ %\cline{2-6}
                    & MLPs        & 67.81 & ${\Delta 0.50}$           & 73.28 & ${\Delta 0.28}$            \\
                    & PwCS        & 70.97 & ${\Delta 3.66}$           & 77.78 & ${\Delta 4.79}$            \\ \cline{2-6}
                    & CRM         & \bf{75.68} & $\Delta\mathbf{8.37}$    & \bf{81.75} & ${\Delta \mathbf{8.76}}$   \\
                    & ~-base      & 69.33 & ${\Delta 2.02}$           & 78.48 & ${\Delta 5.49}$            \\
                    & ~-Xpart     & 66.47 & ${\nabla 0.84}$           & 77.00 & ${\Delta 4.01}$            \\
                    & ~-Xclass    & 74.61 & ${\Delta 7.30}$           & 79.87 & ${\Delta 6.82}$            \\ \midrule
            \multirow{7}{*}{\rotatebox{90}{ViT-B/16}} 
                    & PLOT        & 77.55 & ${\mathrm{Ref.}}$         &  84.42 & ${\mathrm{Ref.}}$         \\ %\cline{2-6}
                    & MLPs        & 76.70 & ${\nabla} 0.85$           &  81.67 & ${\nabla 2.75}$           \\
                    & PwCS        & 77.37 & ${\nabla} 0.18$           &  84.61 & ${\Delta 0.19}$           \\ \cline{2-6}
                    & CRM        & \bf{81.08} & $\Delta\mathbf{3.53}$    &  \bf{86.52} & ${\Delta \mathbf{2.10}}$  \\
                    & ~-base      & 76.60 & ${\nabla 0.95}$           &  84.70 & ${\Delta 0.28}$           \\
                    & ~-Xpart     & 76.42 & ${\nabla 1.13}$           &  85.53 & ${\Delta 1.11}$           \\
                    & ~-Xclass    & 79.11 & ${\Delta 1.56}$           &  86.04 & ${\Delta 1.62}$           \\ 
            \hline 
            \specialrule{0.1em}{1pt}{1pt}
            \end{tabular}
    }
}
\caption{Performance of different prediction strageties.}
\vspace{-1em}
\label{tab:pred}
\end{table}

\subsubsection{Comparison of Different Prediction Strageties.}
\label{sec:pred-strategy}
We propose a cross relationship modeling scheme for prediction, which enables 1-to-many information utilization of prompt and visual features. In this section, we demonstrate its effectiveness for VLM-based FGVR by comparing it with other prediction strategies. The results are presented in Table~\ref{tab:pred}, where \textbf{MLPs} refers to adding an MLP block as the classifier for visual features, and \textbf{PwCS} denotes calculating the similarity between each pair of textual and visual features, as formulated in Eq.~\eqref{eq:pwcs}. For the proposed CRM, as shown in Fig.~\ref{fig:cross}, we evaluate four variants: the base version, Xclass, Xpart, and the full version (Xclass+Xpart). Details of these methods can be found in supplementary material. PLOT is adopted as the baseline since it also interpolates multiple prompts into VLMs. 

We can see that for RN50 visual encoder, PwCS, MLPs, and CRM all achieve better performance than PLOT, and CRM can obtain obvious improvements of 8.37\% and 8.76\%, respectively. As for the ViT encoder, even PLOT achieves better performance than MLPs and PwCS, our proposed CRM still obtains 3.53\% and 2.10\% gains than it. The results demonstrate that: (1) learning multiple prompts can significantly boost the performance on CLIP on fine-grained image classification, but how to utilize it has a significant impact on performance; (2) aligning multiple prompts to multiple visual part features can obtain better performance than to multiple visual local features; (3) the proposed cross relationship modeling module can better explore the correlations between visual and textual information than just part-wise similarity calculating.

\begin{figure}[h]
    \centering
        \begin{subfigure}{0.495\linewidth}
            \includegraphics[width=\textwidth]{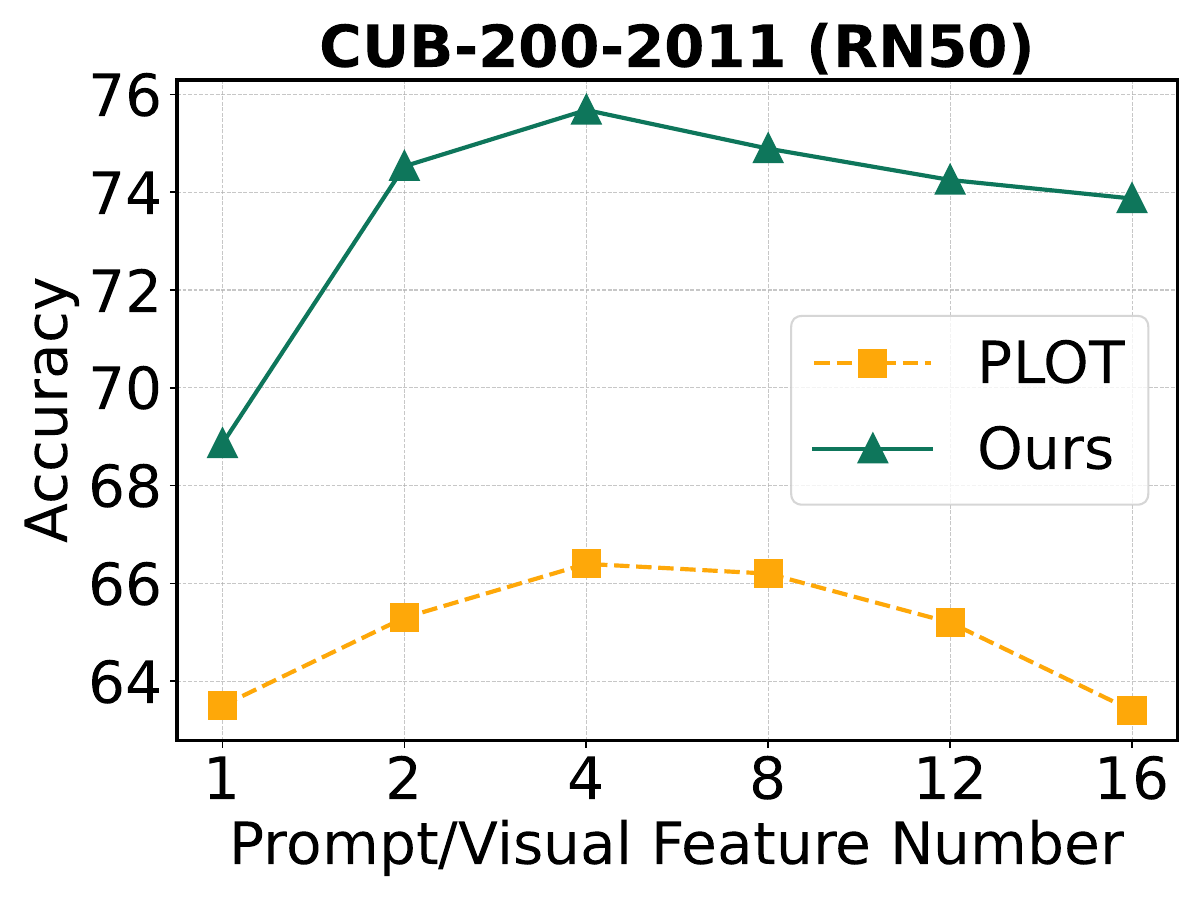}
            % \caption{An example of a subfigure.}
            % \label{fig:short-a}
        \end{subfigure}
    \hfill
        \begin{subfigure}{0.495\linewidth}
            \includegraphics[width=\textwidth]{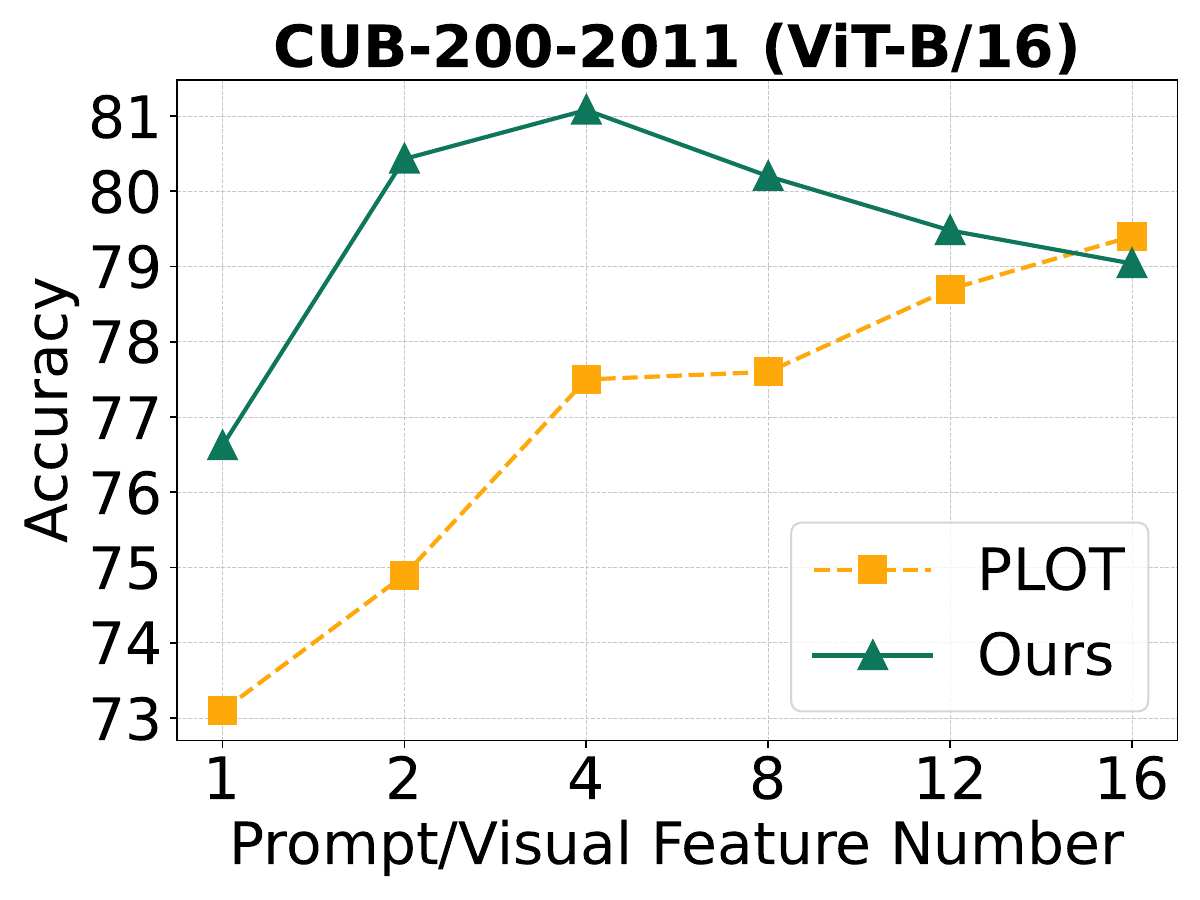}        
            % \caption{Another example of a subfigure.}
            % \label{fig:short-b}
        \end{subfigure}
    
    \begin{subfigure}{0.495\linewidth}
            \includegraphics[width=\textwidth]{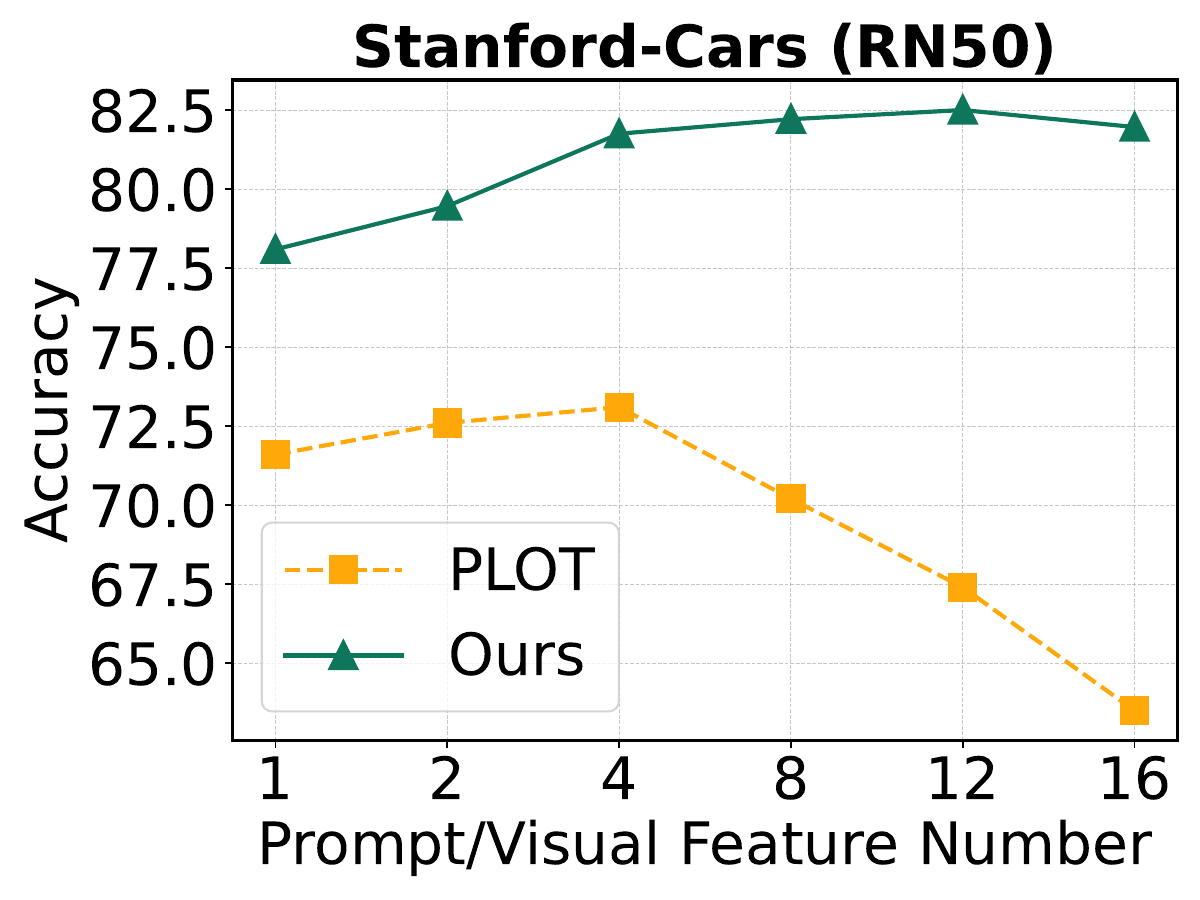}
            % \caption{An example of a subfigure.}
            % \label{fig:short-a}
        \end{subfigure}
    \hfill
        \begin{subfigure}{0.495\linewidth}
            \includegraphics[width=\textwidth]{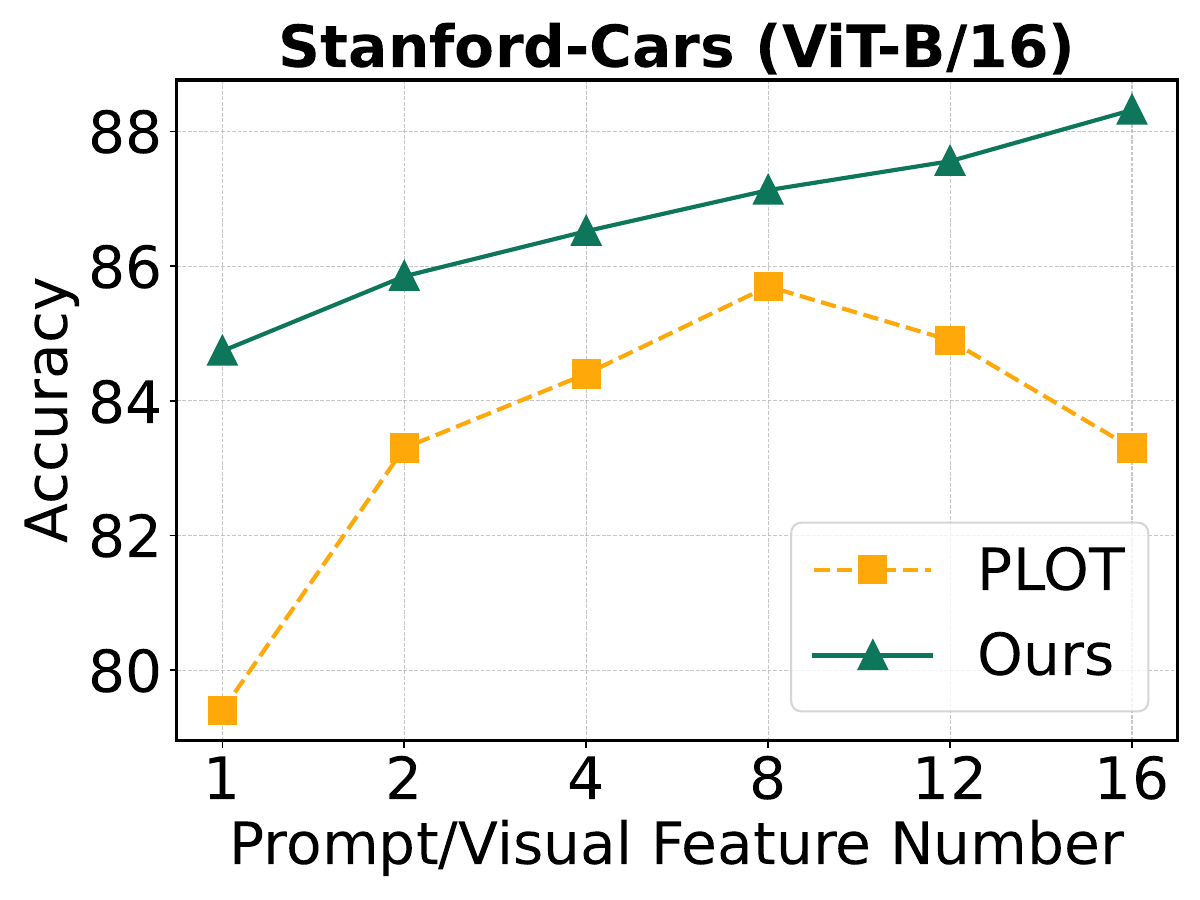}        
            % \caption{Another example of a subfigure.}
            % \label{fig:short-b}
        \end{subfigure}
    \caption{Comparison between PLOT and our method with different number of Prompt/Visual features and image encoders.}
    \label{fig:num_prompts}
    \end{figure}

\begin{figure}
	\centering
	\includegraphics[width=\linewidth]{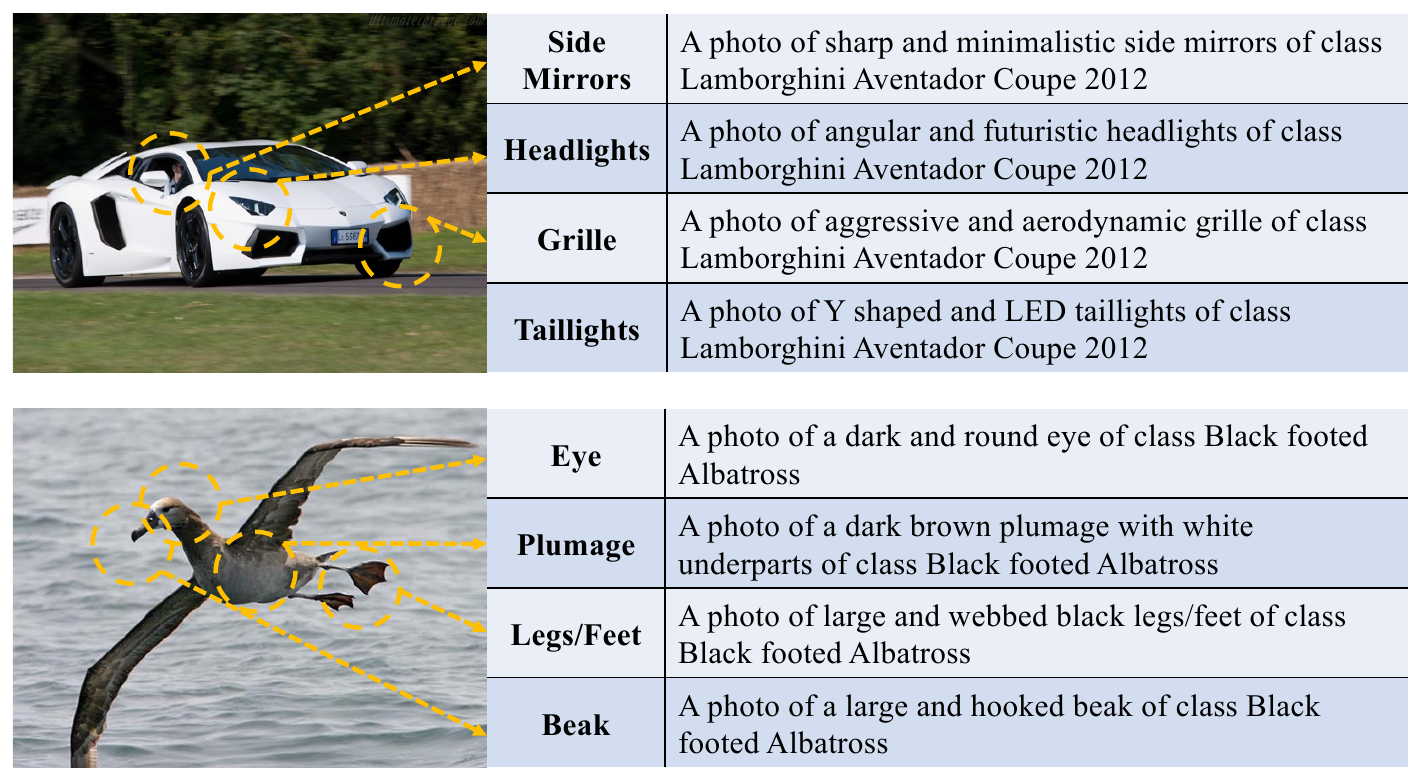}
	\caption{Examples of part descriptions generated by LLM. We let LLM generate 4 key part names of each dataset and corresponding part descriptions of each sub-categories.}
    \vspace{-1em}
	\label{fig:hc-prompt}
\end{figure}

\begin{figure}
	\centering
	\includegraphics[width=\linewidth]{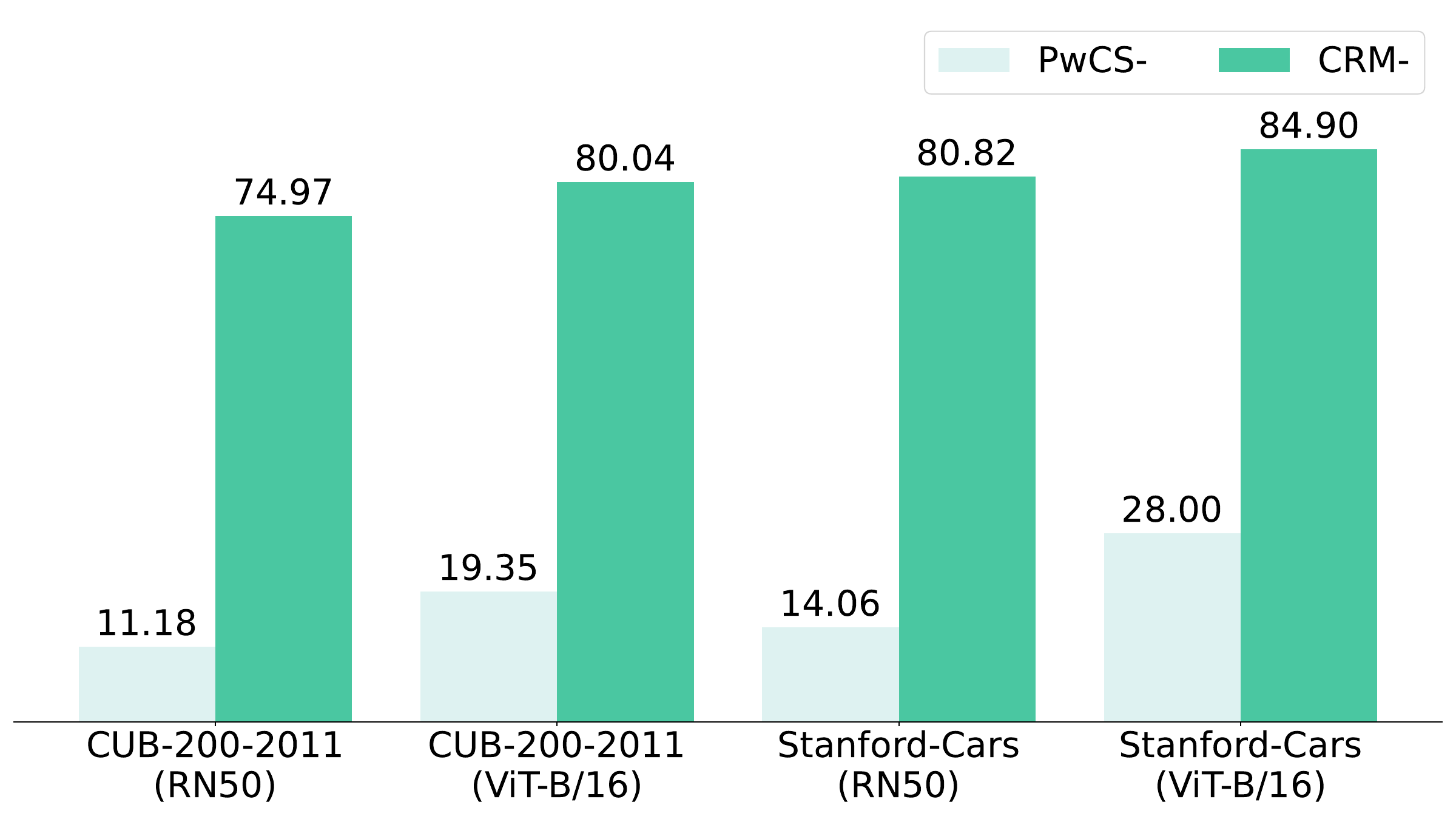}
	\caption{Performance on manual prompts. PwCS- and CRM- denotes modeling relationship between manual prompts and visual featuers by PwCSa and our CRM modules, respectively.}
    \vspace{-1em}
	\label{fig:performance-hc-prompt}

\end{figure}

\begin{figure*}
    \centering
        \begin{subfigure}{0.33\linewidth}
            \includegraphics[width=\textwidth]{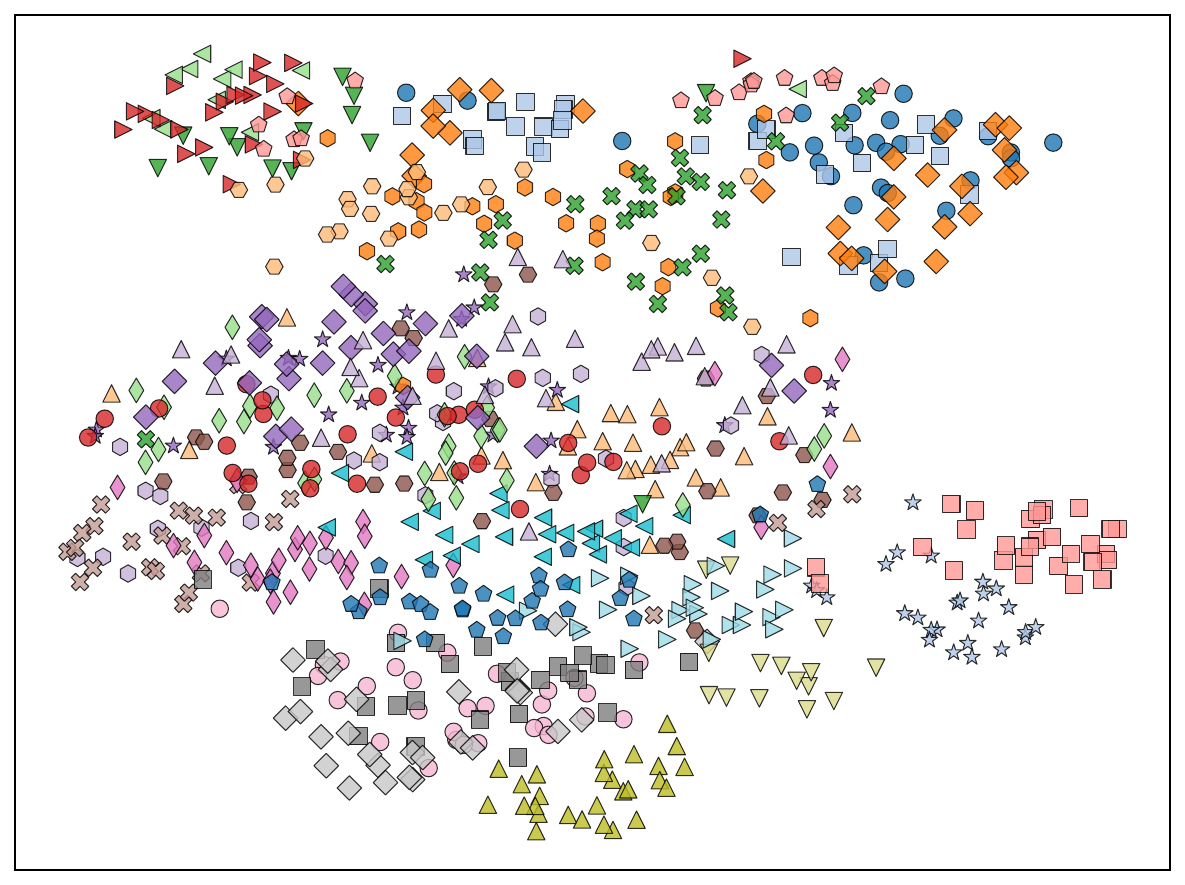}
            \caption{Visual Features Learned by PwCS.}
        \end{subfigure}
    \hfill
        \begin{subfigure}{0.33\linewidth}
            \includegraphics[width=\textwidth]{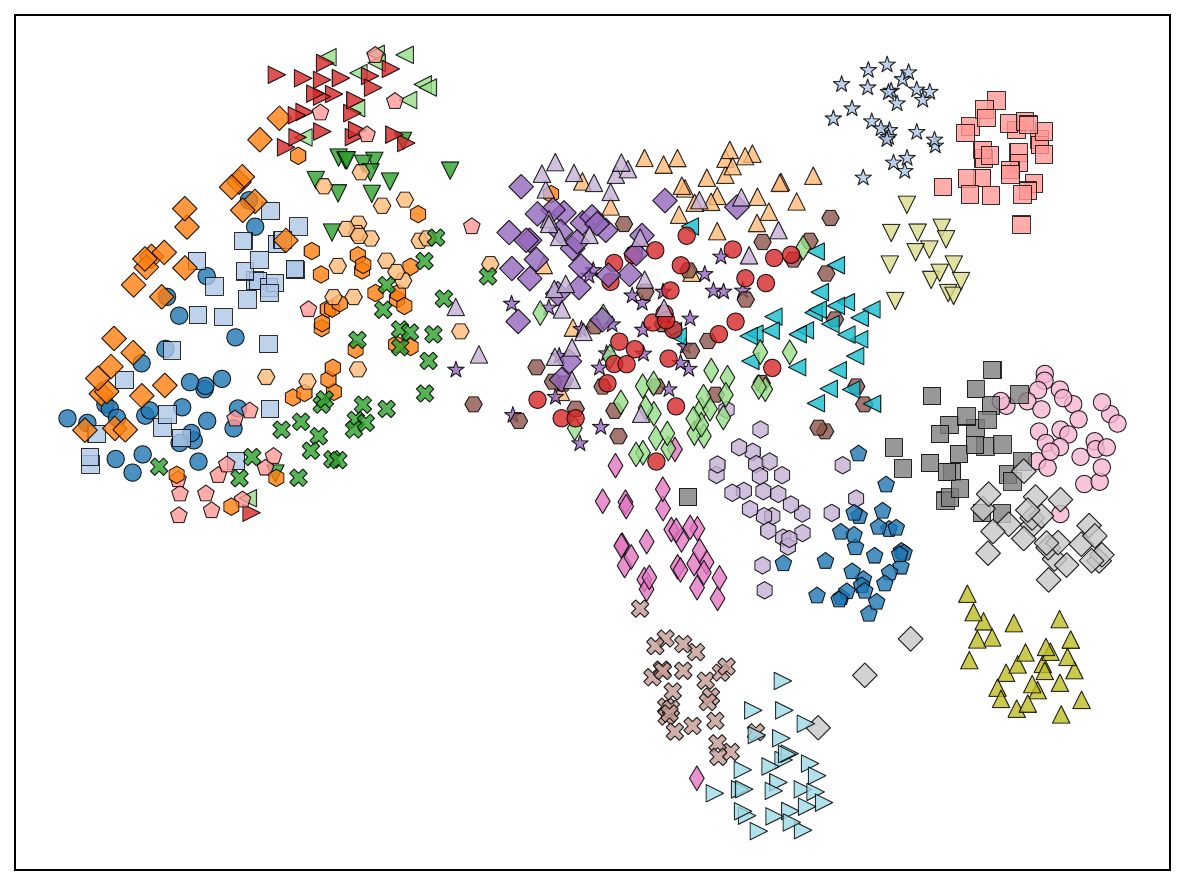}        
            \caption{Visual Features Learned by CRM.}
        \end{subfigure}
    \hfill
        \begin{subfigure}{0.33\linewidth}
            \includegraphics[width=\textwidth]{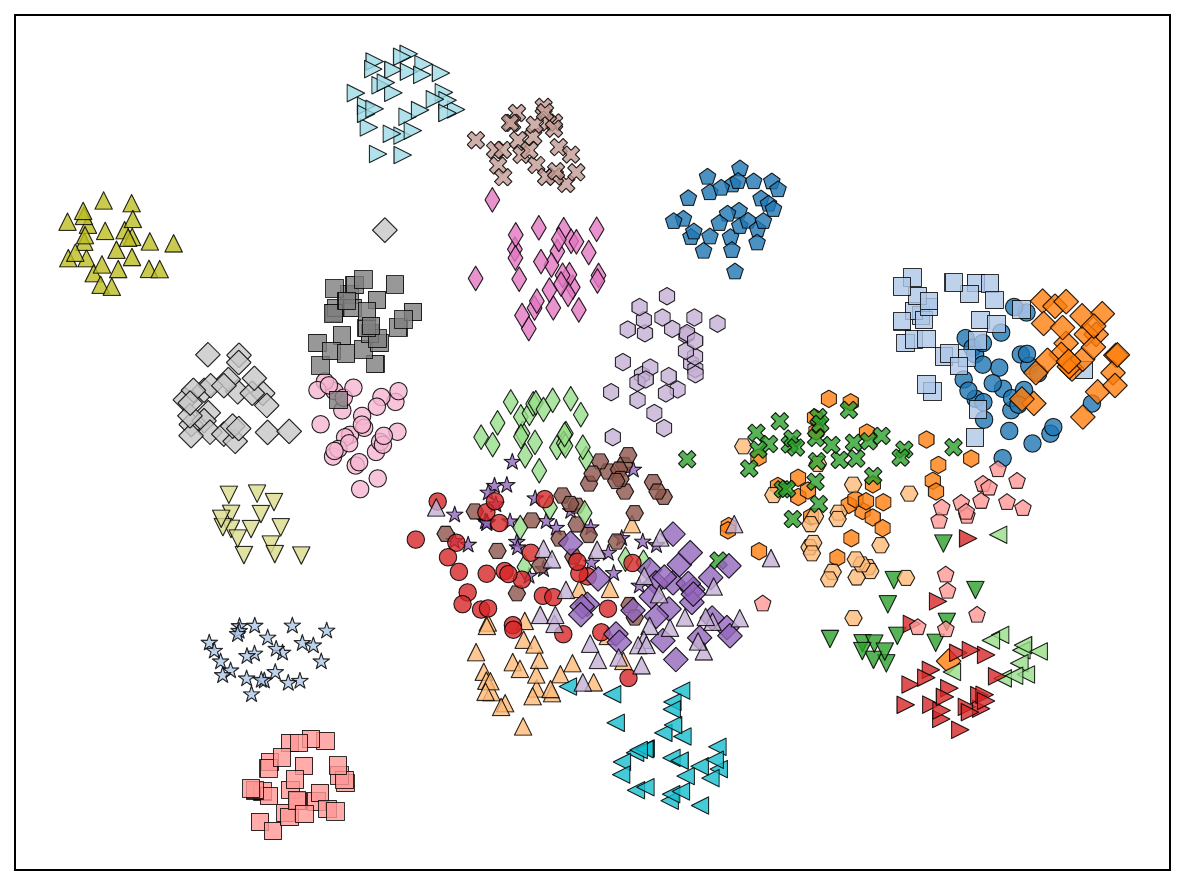}        
            \caption{Cross Relationship Learned by PwCS.}
        \end{subfigure}
    \caption{UMAP visualization of learned features. Different shape-color combinations denote different classes. Data are selected from CUB-200-2011 datasets, and the first 30 classes are shown for clarity of presentation. (best view in color)}
    \label{fig:umap}
    \end{figure*}

\subsubsection{Different Number of Prompts/Visual Features.}
Our proposed XR-VLM processes multiple prompts and visual features for prediction. To assess the impact of varying numbers of prompts and visual features, we conduct experiments and compare our method with PLOT. The results are presented in Fig.~\ref{fig:num_prompts}. As the number of prompt/visual representations increases, our method consistently outperforms PLOT across most of settings, demonstrating superior adaptability and robustness in fine-grained recognition. Specifically, (1) our method usually achieves significantly higher accuracy than PLOT across different dataset-backbone combinations. This indicates that modeling crossing relationship can effectively enhances fine-grained recognition; (2) for the ViT-B/16 backbone, while PLOT exhibits performance degradation as the number of prompts/visual representations increases, our method maintains stable or improving trends. Notably, although a larger number of prompts/visual features generally leads to better performance, an excessive number introduces more learnable parameters, which can slow down the inference speed of the text encoder. To balance performance and efficiency, we set the number to 4 as default.

\subsubsection{Compatible with Manual Prompts.}
\label{sec:mp}
In this work, we propose a novel prediction pattern that takes a further step in leveraging both prompt and visual features for fine-grained recognition. Our method not only integrates these two modalities but also treats their interaction as a discriminative representation. By default, multi-part prompts are learned through back-propagation, ensuring adaptive optimization, while this pattern also supports manually designed prompts for greater flexibility. To achieve this, we utilize a large language model, DeepSeek-V3, to generate a set of part-level descriptions for the CUB-200-2011 and Stanford-Cars datasets as the prompts. Some examples are shonw in Fig.\ref{fig:hc-prompt}. By comparing the images with their corresponding descriptions, we observe that some generated descriptions can effectively capture fine-grained class-specific details, but there are some words that struggle to express category-specific features, \eg, `aggressive and aerodynamic grille'. The results are shown in Fig.~\ref{fig:performance-hc-prompt}. We can find that PwCS is hard to handle these manual prompts but our method can still achieve high perofmance, just a little lower than using learnable prompts.

\subsection{Visualization}

\noindent \textbf{Features Visualization.} To demonstrate the discriminative power of cross-relationship modeling, we visualize the visual features and cross relationships by UMAP~\cite{DBLP:journals/corr/abs-1802-03426} technology in Fig.~\ref{fig:umap}. As shown in the figure, the visual features from the PwCS are difficult to distinguish on the 2D plane, and while some classes from CRM exhibit clustering, the separation is still not pronounced. In contrast, the cross relationship of each class is relatively distinct, highlighting its discriminative capability and explaining the significant performance improvement observed in our experiments.

\begin{figure}
    \centering
    \includegraphics[width=\linewidth]{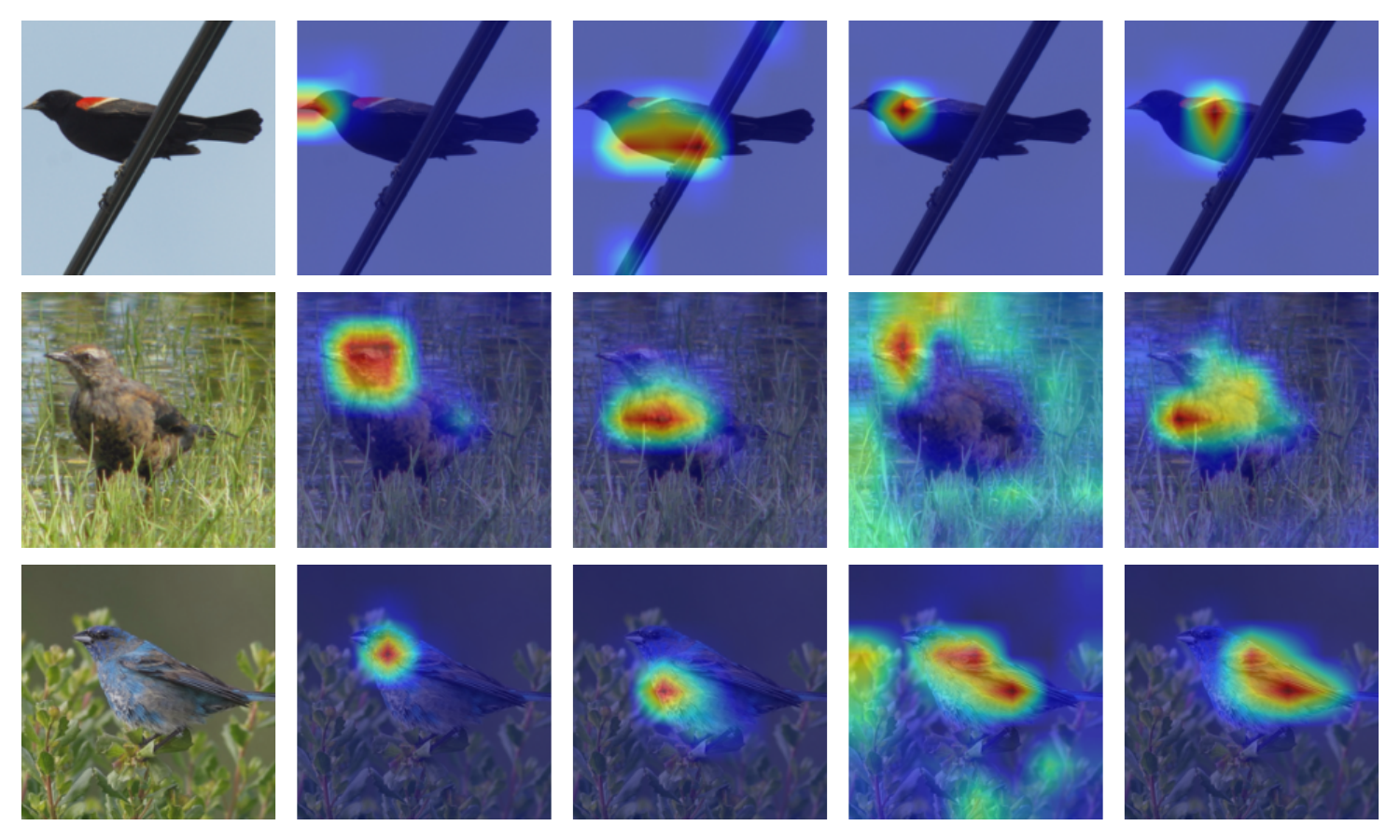}
    \caption{Visualization of learned attentions.}
    \label{fig:atten}
    \vspace{-1em}
\end{figure}

\noindent \textbf{Attention Visualization.} 
Then, we visualize the generated attention masks in Fig.~\ref{fig:atten} to provide insights into the characteristics of the visual features. As shown in the figure, for each input image (first column), the generated attention masks effectively cover different parts of the birds. This demonstrates that the unified attention module successfully extracts discriminative multi-part visual features, thereby enhancing the effectiveness of cross-relationship modeling.

%% file: sec/5_conclusion.tex
\section{Conclusion}

In this paper, we find that current fine-tuning methods for VLMs like CLIP face challenges to the task of fine-grained visual recognition. We argue that this limitation stems from restricted prompt/feature learning and insufficient utilization of relationships between modalities. To address this, we propose a multi-part prompt learning and utilization method that learns both multiple textual and visual features and deeply leverages their relationships for FGVR. Experimental results demonstrate that our method achieves state-of-the-art performance across various benchmark fine-grained datasets, outperforming a wide range of existing methods. 
We should acknowledge that current method does not take the domain generalization into considierasion and we think it is not in the scope of this work, and we will focues on how to model adptive relationship for incremental learning and base-to-novel learning in the future. 

%% file: sec/X_suppl.tex
\clearpage
\setcounter{page}{1}
\maketitlesupplementary
\begin{figure*}[!t]
    \centering
    \includegraphics[width=0.9\linewidth]{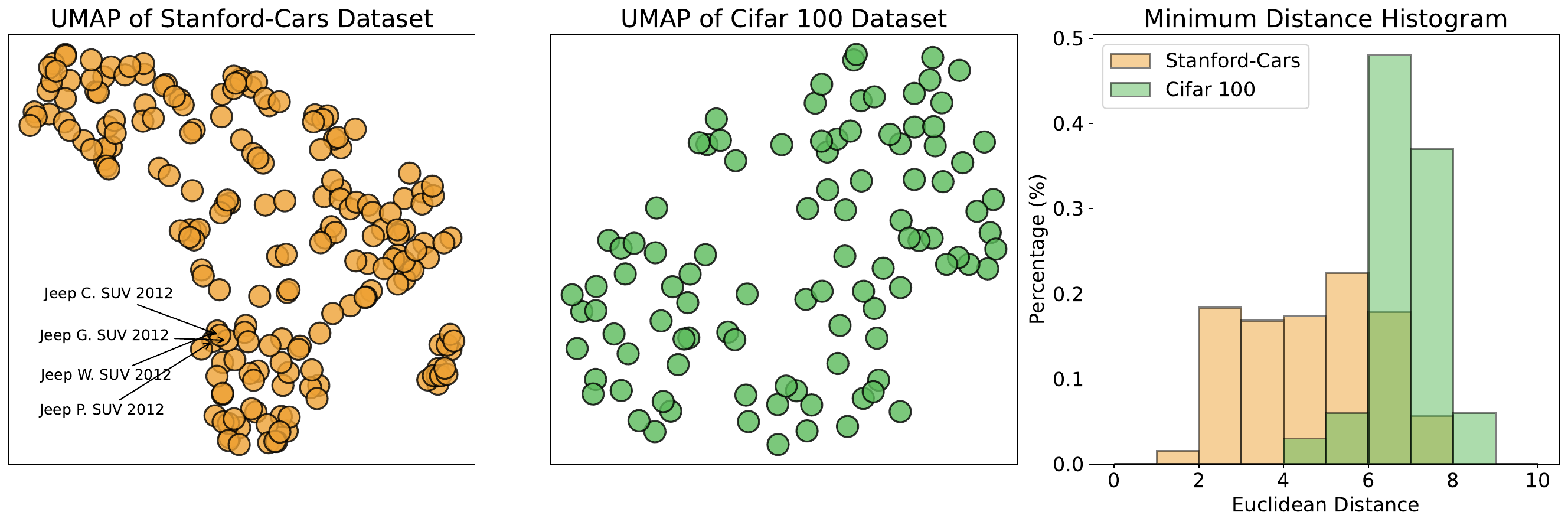}
    \vspace{-1em}
    \caption{UMAP of class name embddings and minimum distance histogram.}
    \label{fig:scatter-plot}
\end{figure*}

\section{Similarity between Fine-grained Classes}

Fine-grained visual recognition focuses on distinguishing sub-categories within a broader superclass, where the similarity among these sub-categories is significantly higher compared to coarse-grained classes, often leading to confusion in predictions. The high similarity between class names also poses a challenge for VLMs, which rely on class names as the basis for classification, making it difficult to obtain discriminative features. In this section, we demonstrate this through a simple experiment and visualization. We evaluate the Stanford-Cars and CIFAR-100 datasets, using the ViT-B/16 text encoder from CLIP. All class names from both datasets are encoded by the text encoder, and we apply UMAP to visualize their embeddings.

As shown in Fig.~\ref{fig:scatter-plot} (in next page), from the visualization, we observe that the Class Name Embeddings (CNEs) in CIFAR-100 are more evenly distributed, whereas those in Stanford-Cars are closer to each other in the embedding space. This indicates that distinguishing fine-grained categories based solely on category names is more challenging for Stanford-Cars, as the semantic differences between class names are less pronounced. In the first column of Fig.~\ref{fig:scatter-plot}, we highlight some class names that are close to each other, showing that their names are highly similar, differing only in model types. Consequently, a simple alignment-based prediction method would struggle to learn discriminative features for fine-grained recognition in such scenarios.

To further quantitatively analyze the similarity of CNEs across different datasets, we calculate the distance between each CNE and its nearest neighbor and plot the corresponding histogram (third column of Fig.~\ref{fig:scatter-plot}). The histogram for CIFAR-100 is relatively flat, with most distances being less than 7. In contrast, for Stanford-Cars, most minimum distances are greater than 6, indicating that the class names are much more similar.

\section{Details of Unified Attention Module}

Since there are two architectures (ResNet and Vision Transformer) of visual encoder in CLIP, and their outputs are different in shapes. In Section~\ref{sec:uni-atten}, we introduce an unified attention module to extract multi-part visual features that can fit with these two types of architectures. Thus, in this part, we detail how it works for different architectures.

For ResNet in CLIP, it consists of one input term block (Conv-BN-ReLU-Maxpool), four residual layers, and one attention pooling layer. The attention pooling layer is implemented by a multihead attention module, and it can be formulated as:
\begin{equation}
    \begin{aligned}
        \mathbf{F} =& \operatorname{softmax}\left(\frac{\mathbf{\bar{\mathbf{X}}}\mathbf{W}_q(\mathbf{\bar{\mathbf{X}}}\mathbf{W}_k)^\top}{\sqrt{d_k}}\right) (\mathbf{\bar{\mathbf{X}}}\cdot \mathbf{W}_v) \mathbf{W}_c, \\
        \mathbf{\bar{X}} =& [\operatorname{AVG}_{2D}(\mathbf{X}), \epsilon \left(\mathbf{X}, (HW\times C)\right)] + \mathbf{P}.
    \end{aligned}
    \label{eq:res-ap}
\end{equation}
In Eq.~\eqref{eq:res-ap}, $\mathbf{X} \in \mathbb{R}^{C\times H\times W}$ is the output of residual layers and input of attention pooling layer, $\operatorname{AVG}_{2D}$ denotes the average pooling layer for the spatial dimension, $\epsilon$ is the reshape operation that re-arrange the shape of $\mathbf{X}$ to $HW \times C$, and $\mathbf{P} \in \mathbb{R}^{(1+HW) \times C}$ is the position embeddings. Then the self-attention operation is performed on $\bar{\mathbf{X}}$ and outputs $\mathbf{F} \in \mathbb{R}^{(1+HW) \times C}$. In the original CLIP, the first feature in $\mathbf{F}$ is selected as the output of the attention pooling layer, but it losses too many spatial information for part feature learning. Thus, keeping the parameters fixed, we simplify the visual feature learning process as:
\begin{equation}
    \begin{aligned}
        \mathbf{F} = \epsilon\left(\left(\mathbf{\bar{\mathbf{X}}}\mathbf{W}_v\right) \mathbf{W_c}, HW\times C\right) \in \mathbb{R}^{HW\times C}.
    \end{aligned}
\end{equation}
Then, $\mathbf{F}$ is sent to our proposed unified attention module for multi-part visual feature learning.

As for Vision Transofrmer, it outputs $\mathbf{F}$ with shape ${N\times C}$ as the output of the image encoder, we directly use it as the input of our unified attention module for multi-part visual feature learning.

\section{Details of Prediction Strategy.}
In Section~\ref{sec:pred-strategy} of the main text, we introduce various prediction strategies to demonstrate that our proposed CRM can achieve optimal performance. In addition to the \textbf{PwCS} introduced in Eq.~\eqref{eq:pwcs}, we will describe the remaining prediction strategies in this part.

\noindent \textbf{MLPs}. To maintain consistency with the symbols used in the main text, we denote the multi-part visual features as $\mathbf{V} \in \mathbb{R}^{S\times E}$, where $S$ represents the number of parts and $E$ is the feature dimension. We employ $N$ MLPs, denoted as $\{f_1, \dots, f_N \}$, as classifiers, and their outputs are averaged to produce the final prediction:
\begin{equation}
    \mathbf{\hat{y}} = \frac{1}{N}\sum_i^S f_i(\mathbf{V}_i).
\end{equation}
Then, the loss is computed using the cross-entropy loss function between the predicted output $\mathbf{\hat{y}}$ and the ground-truth labels $\mathbf{y}$, optimizing the parameters in prompts, the Unified Attention module, and the MLPs.
Although no language information is used in this prediction, the results are still comparable to those of CoOp on the CUB-200-2011 and Stanford-Cars datasets. This indicates that learning multi-part visual features is beneficial for VLM-based FGVR. However, its performance remains significantly lower than that of the CRM strategy, underscoring the importance of leveraging both visual and language information in VLMs.

\begin{figure}[t]
	\centering
	\includegraphics[width=\linewidth]{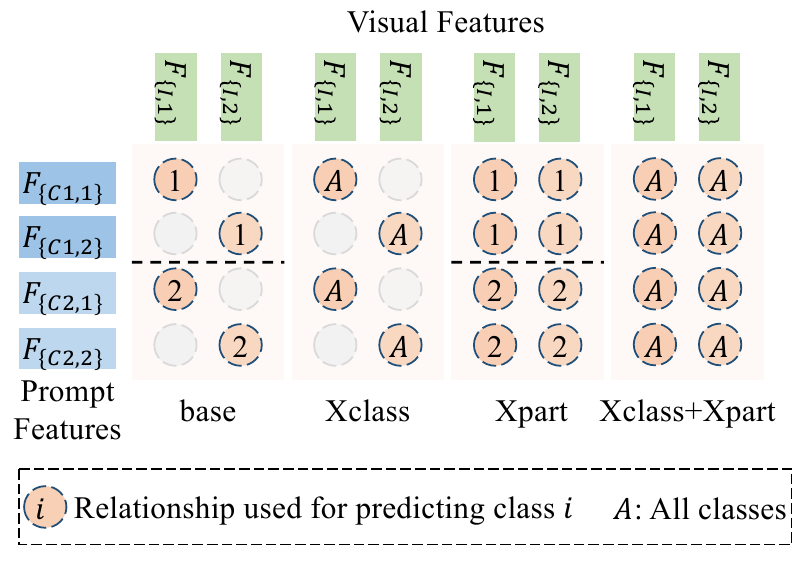}
    \vspace{-1.5em}
	\caption{Illustration of different cross relationships for prediction.}
	\label{fig:cross-sm}
    \vspace{-1em}
\end{figure}

\noindent \textbf{CRM-base}. Let the learned cross relationship from Eq.~\eqref{eq:ips-sim} be denoted as $\mathbf{R}\in \mathbb{R}^{SSW}$, where $W$ is the number of classes in the current dataset. We first reshape $\mathbf{R}$ to the shape of $W\times S\times S$. Then, we pass it through an MLP layer $f$ (output dimension is 1) to obtain the final prediction, formulated as:
\begin{equation}
   \mathbf{\hat{y}} = \epsilon\left([f(\epsilon(\mathbf{R}_1 \mathbf{I}_S)), \cdots, f(\epsilon(\mathbf{R}_W \mathbf{I}_S))]\right) \in \mathbb{R}^{W},
\end{equation}
where $\epsilon$ denotes the flatten operation, $\mathbf{I}^S$ is the identity matrix, and $[\cdots]$ represents the concatenation operation.

For clarity, we include Fig.~\ref{fig:cross} of the main text here as Fig.~\ref{fig:cross-sm}. By using the identity matrix, the cross-part relationships are masked, and the relationship with number 1 is used to predict the probability of class 1, and so on.

\noindent \textbf{CRM-Xclass}. Same with CRM-base, we employ $\mathbf{R}$ to denote the learned cross relationships, and we still adopt a MLP to output the prediction, but its output dimension is $W$. It can be formulated as:
\begin{equation}
    \mathbf{\hat{y}} = f([\mathbf{R}_1\mathbf{I}_S, \cdots, \mathbf{R}_W\mathbf{I}_S])\in \mathbb{R}^{W}.
\end{equation}

\noindent \textbf{CRM-Xpart}. Still denote $\mathbf{R}$ the cross relationships, the prediction of CRM-Xclass can be formulated as:
\begin{equation}
    \mathbf{\hat{y}} = \epsilon\left([f(\epsilon(\mathbf{R}_1)), \cdots, f(\epsilon(\mathbf{R}_W))]\right) \in \mathbb{R}^{W},
 \end{equation}
where $f$ is a MLP, and its output dimension is 1.

\section{Manual Prompts}

In Section~\ref{sec:mp}, we introduce manual prompts to demonstrate the robustness of our proposed Cross-Relationship Modeling to different prompts. We detail the process of generating these prompts and provide additional examples in this section.

\begin{table*}[]
    \begin{tabularx}{\textwidth}{cX}
    \hline
    \textbf{Roles} & \multicolumn{1}{c}{\textbf{Content}}                                                             \\ \hline
    User  & I am doing fine-grained image classification, can you provide the names of 4 key parts to distinguish different cars? Just give me the answer.                                              \\ \hline
    LLM   & [headlights, grille, side mirrors, taillights]                              \\ \hline
    User &
      \begin{tabular}[c]{@{}l@{}}The user is performing fine-grained image classification and will provide a car class name along with the names \\of a set of key parts. You need to do the following:\\ Based on the provided parts, describe their appearance for the given car class, avoiding abstract words.\end{tabular} \\ \hline
    User  & Lamborghini Aventador Coupe 2012,~[headlights, grille, side mirrors, taillights] \\ \hline
    LLM &
      \begin{tabular}[c]{@{}l@{}}headlights: a photo of angular and futuristic headlights of class Lamborghini Aventador Coupe 2012,\\ grille: a photo of aggressive and aerodynamic grille of class Lamborghini Aventador Coupe 2012,\\ side mirrors: a photo of sharp and minimalistic side mirrors of class Lamborghini Aventador Coupe 2012,\\ taillights: a photo of Y shaped and LED taillights of class Lamborghini Aventador Coupe 2012.\end{tabular} \\ \hline
    \end{tabularx}
    \caption{Instructions for LLM to generate part-level descriptions.}
    \label{tab:inp-prompt}
    \end{table*}

\begin{figure*}[h]
    \centering
        \begin{subfigure}{0.495\linewidth}
            \includegraphics[width=\textwidth]{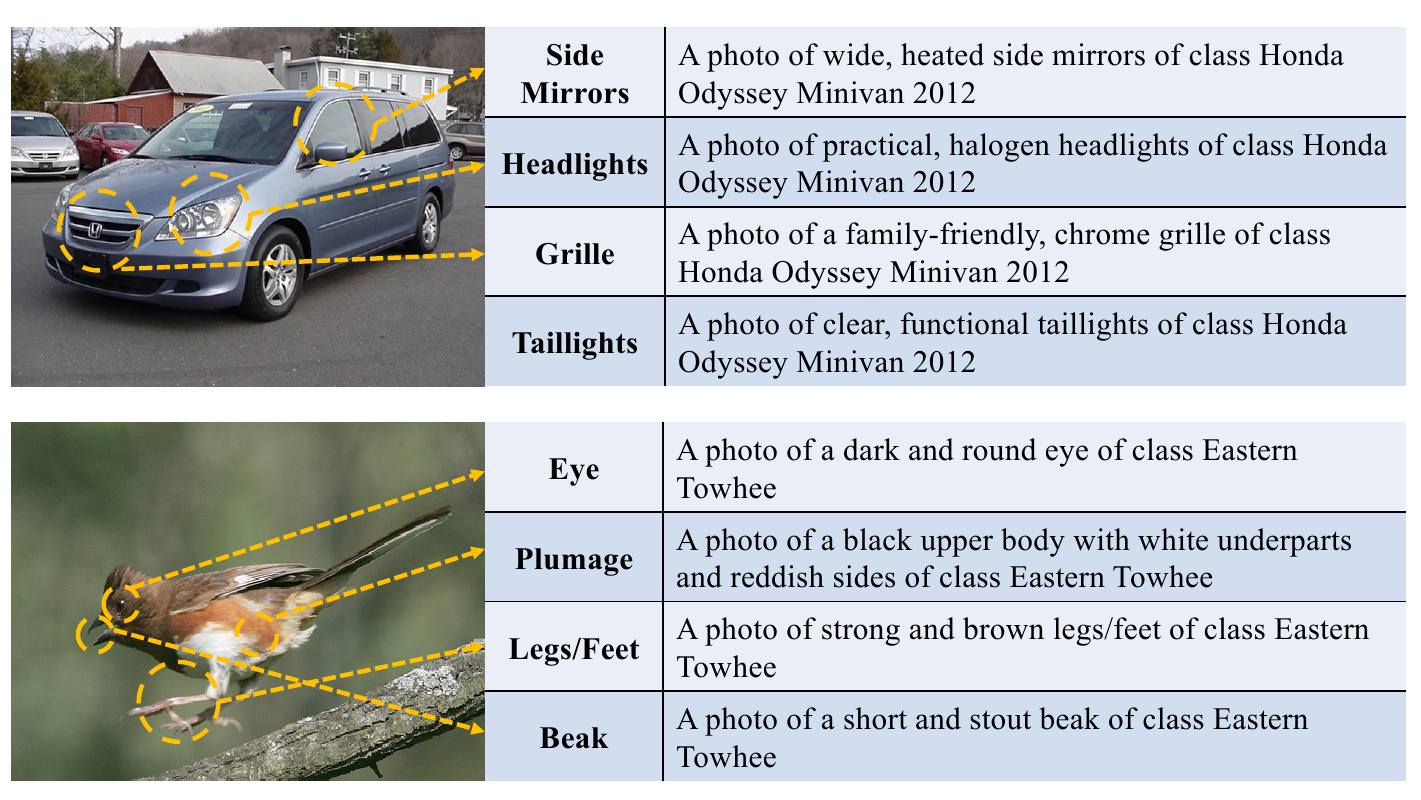}
            % \caption{An example of a subfigure.}
            % \label{fig:short-a}
        \end{subfigure}
    \hfill
        \begin{subfigure}{0.495\linewidth}
            \includegraphics[width=\textwidth]{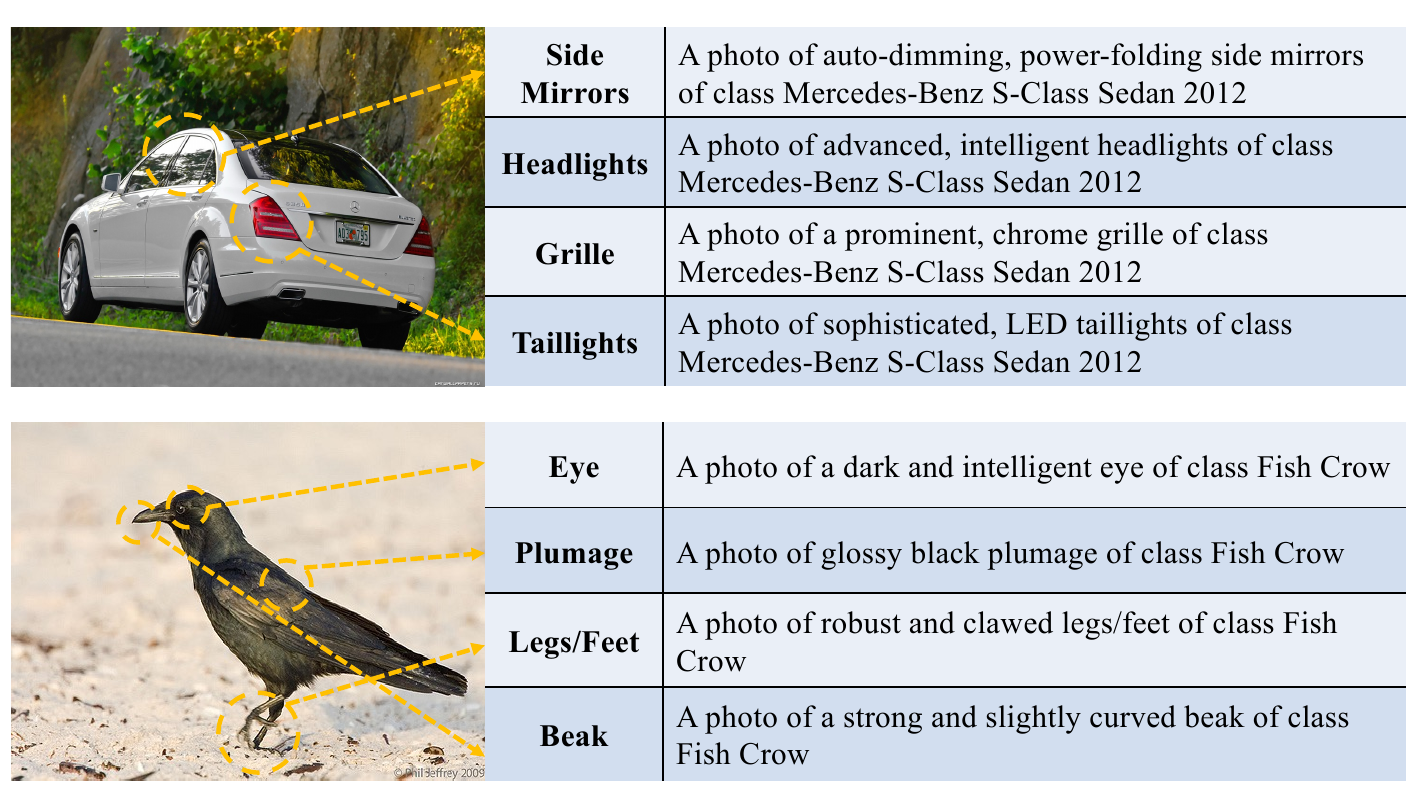}        
            % \caption{Another example of a subfigure.}
            % \label{fig:short-b}
        \end{subfigure}
    
    \begin{subfigure}{0.495\linewidth}
            \includegraphics[width=\textwidth]{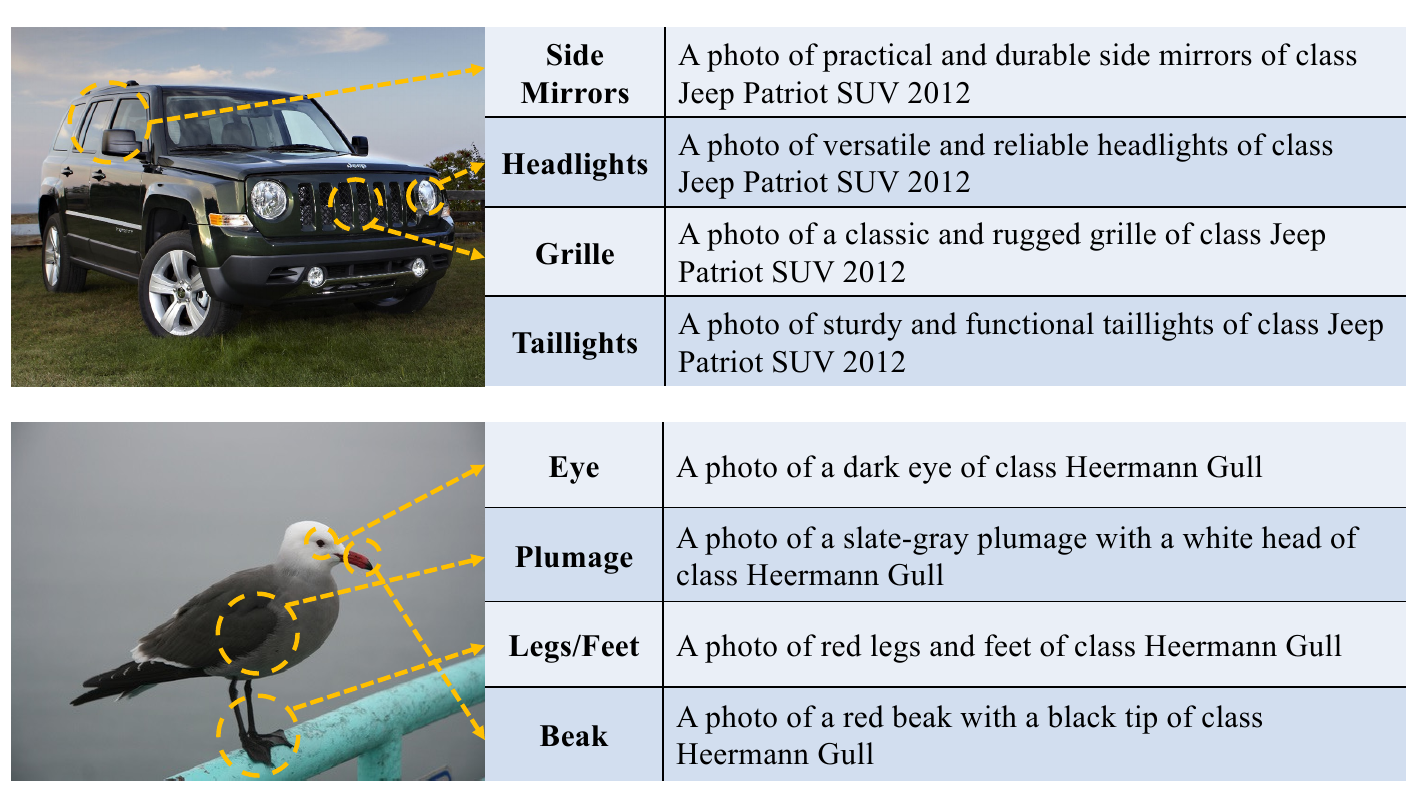}
            % \caption{An example of a subfigure.}
            % \label{fig:short-a}
        \end{subfigure}
    \hfill
        \begin{subfigure}{0.495\linewidth}
            \includegraphics[width=\textwidth]{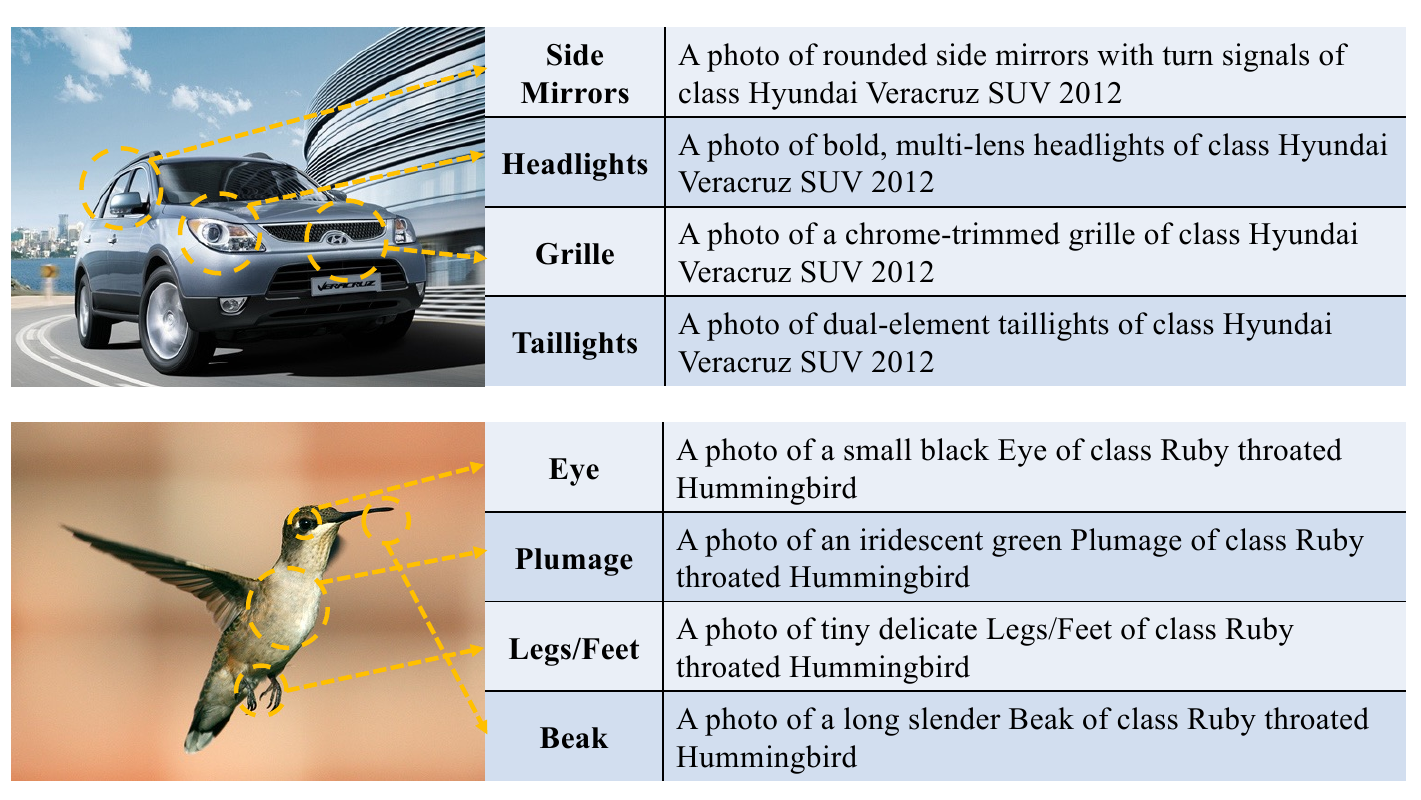}        
            % \caption{Another example of a subfigure.}
            % \label{fig:short-b}
        \end{subfigure}
    \caption{Additional examples of manual prompts for CUB-200-2011 and Stanford-Cars datasets.}
    \label{fig:more-mp}
    \end{figure*}

Table~\ref{tab:inp-prompt} outlines the instructions used to generate the manual prompts. We first instruct a LLM, specifically DeepSeek-V3 in our experiments, to output four key part names that can be used to distinguish different sub-categories. For example, it generates `side mirrors', `headlights', `grille', and `taillights' for cars, and `eye', `beak', `legs/feet', and `plumage' for birds. Next, we provide the LLM with a class name and ask it to generate descriptions of that class based on the previously identified parts.

In Fig.~\ref{fig:more-mp}, we present additional manual prompts. It is evident that the quality of the generated descriptions is inconsistent, with some descriptions being overly vague and failing to accurately capture the characteristics of the target class. We speculate that this issue may arise from two factors: (1) LLM may require more fine-tuned instructions to produce higher-quality part descriptions, and (2) describing the characteristics of fine-grained sub-categories in explicit language may be inherently challenging, even for advanced language models.

\section{Computational Overhead}

We acknowledge that multi-part prompt/feature learning introduces additional parameters, which may increase inference time. Here, we provide a detailed analysis. When the number of prompts/visual features is set to 4 and the image encoder is RN50, the trainable parameters for the CUB-200-2011 dataset total 11.1M, distributed as follows: prompts account for 6.6M parameters, the Unified Attention module contains 14.3K parameters, and the classifier comprises 4.5M parameters. Compared to the original CLIP, which has 76.7M parameters, the trainable parameters in our method are relatively small, representing only 12.6\% of total parameters. For ViT-B/16, this percentage decreases to 7.7\%.

Additionally, we evaluate the inference time of our method against CoOp. The results show that CoOp and our method takes 13.4 ms and 15.3 ms for RN50, and 21.95 ms and 41.8 ms for ViT-B/16, respectively. In future work, we will explore ways to optimize inference speed and reduce the number of learnable parameters by matrix decomposition, q-former or other strategies.